\documentclass[journal]{IEEEtran}
\usepackage{graphicx}
\usepackage{amsmath, amssymb}
\usepackage{soul, xcolor}
\usepackage{caption, subcaption}
\usepackage{hyperref}
\usepackage{cite}
\usepackage{url}
\usepackage{footnote}
\makesavenoteenv{tabular}
\usepackage{multirow}
\usepackage{tikz}
\usetikzlibrary{positioning,calc,arrows.meta,decorations.pathreplacing}
\usepackage{xcolor}
\definecolor{gold}{RGB}{220,0,0}      
\definecolor{silver}{RGB}{0,160,0}    
\definecolor{bronze}{RGB}{0,90,255}   
\newcommand{\gold}[1]{\textbf{\textcolor{gold}{#1}}}      
\newcommand{\silver}[1]{\textbf{\textcolor{silver}{#1}}}  
\newcommand{\bronze}[1]{\textbf{\textcolor{bronze}{#1}}}  
\usepackage{algorithm}
\usepackage{algorithmic}

\usepackage{tikz}
\usetikzlibrary{arrows.meta, shapes, positioning, calc, fit}
\usepackage{pgfplots}
\pgfplotsset{compat=1.18}
\usepackage{pgfplotstable}
\usetikzlibrary{patterns}
\begin{document}
\title{Cosine-Normalized Attention for Hyperspectral Image Classification}
\author{Muhammad Ahmad, Manuel Mazzara 
\thanks{M. Ahmad is with SDAIA-KFUPM, Joint Research Center for Artificial Intelligence (JRCAI), King Fahd University of Petroleum and Minerals, Dhahran, 31261, Saudi Arabia. (e-mail: mahmad00@gmail.com).}
\thanks{M. Mazzara is with the Institute of Software Development and Engineering, Innopolis University, Innopolis, 420500, Russia.}
}
\markboth{ArXiv}
{Ahmad \MakeLowercase{\textit{et al.}}}
\maketitle
\begin{abstract}
Transformer-based methods have improved hyperspectral image classification (HSIC) by modeling long-range spatial–spectral dependencies; however, their attention mechanisms typically rely on dot-product similarity, which mixes feature magnitude and orientation and may be suboptimal for hyperspectral data. This work revisits attention scoring from a geometric perspective and introduces a cosine-normalized attention formulation that aligns similarity computation with the angular structure of hyperspectral signatures. By projecting query and key embeddings onto a unit hypersphere and applying a squared cosine similarity, the proposed method emphasizes angular relationships while reducing sensitivity to magnitude variations. The formulation is integrated into a spatial–spectral Transformer and evaluated under extremely limited supervision. Experiments on three benchmark datasets demonstrate that the proposed approach consistently achieves top-tier performance, outperforming several recent Transformer- and Mamba-based models despite using a lightweight backbone. In addition, a controlled analysis of multiple attention score functions shows that cosine-based scoring provides a reliable inductive bias for hyperspectral representation learning.
\end{abstract}
\IEEEpeerreviewmaketitle

\section{Introduction}

\IEEEPARstart{H}{yperspectral imaging} (HSI) acquires dense spectral signatures for each spatial location in a scene, typically spanning tens to hundreds of contiguous spectral bands \cite{10035973}. Owing to this fine spectral resolution, HSI enables detailed characterization of surface materials and supports a wide range of remote sensing applications, including land-cover mapping, environmental monitoring, precision agriculture, and mineral exploration \cite{AHMAD2025130428,AASEN2018374}. Among these tasks, pixel-wise HSI classification (HSIC) remains one of the most fundamental and widely studied problems, where each pixel is assigned a semantic category based on its spectral and spatial context \cite{11266884,10767233}. Despite substantial progress over the past years, HSIC continues to be challenging because hyperspectral data are high dimensional, exhibit strong inter-band redundancy, and are often accompanied by only a limited number of labeled samples \cite{11431215,9645266}.

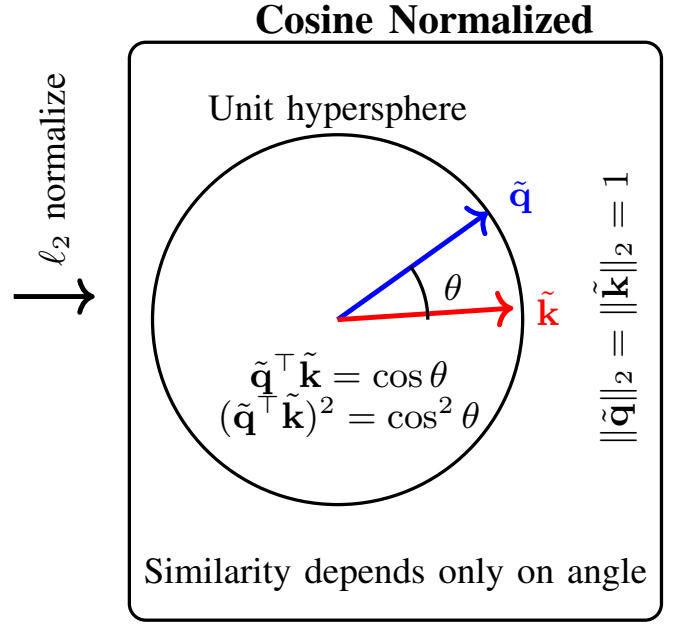
\begin{figure}[!t]
\centering
\resizebox{\columnwidth}{!}{
\begin{tikzpicture}[scale=0.99, every node/.style={font=\small}]
\draw[->, very thick] (4.6,1.1) -- (5.3,1.1);
\node[align=center, rotate=90] at (4.95,2.20) {$\ell_2$ normalize};
\draw[rounded corners, thick] (5.6,-1.7) rectangle (10.2,3.3);
\node[font=\bfseries] at (8.15,3.5) {Cosine Normalized};
\coordinate (C) at (7.4,0.90);
\draw[thick] (C) circle (1.6);
\node at (7.4,2.70) {Unit hypersphere};
\draw[very thick,blue,->] (C) -- ++(1.30,0.93);
\draw[very thick,red,->] (C) -- ++(1.52,0.10);
\draw[thick] ($(C)+(0.78,0)$) arc[start angle=0,end angle=36,radius=0.78];
\node at ($(C)+(1.0,0.28)$) {$\theta$};
\node[blue] at ($(C)+(1.58,1.06)$) {$\tilde{\mathbf{q}}$};
\node[red] at ($(C)+(1.82,0.10)$) {$\tilde{\mathbf{k}}$};
\node[align=center] at (7.90,-1.30)
{Similarity depends only on angle};
\node[align=center] at (7.50,0.25)
{$\tilde{\mathbf{q}}^{\top}\tilde{\mathbf{k}}=\cos\theta$ \\
 $(\tilde{\mathbf{q}}^{\top}\tilde{\mathbf{k}})^2=\cos^2\theta$};
\node[align=center, rotate=90] at (9.80,1.00)
{$\|\tilde{\mathbf{q}}\|_2=\|\tilde{\mathbf{k}}\|_2=1$};
\end{tikzpicture}}
\caption{Standard dot-product attention mixes vector magnitude and angular alignment through $\mathbf{q}^{\top}\mathbf{k}$. While after $\ell_2$ normalization, the embeddings lie on a unit hypersphere and the score depends only on angular similarity. The proposed squared cosine score further sharpens the distinction between strongly aligned and weakly aligned token pairs.}
\label{fig:geometry_attention}
\end{figure}

Deep learning methods, particularly convolutional neural networks (CNNs), have significantly advanced HSIC by learning hierarchical spectral-spatial representations directly from data \cite{11373109,9307220,11359294}. Their success stems from the ability to jointly exploit local neighborhood structure and spectral variation. Nevertheless, CNNs are inherently constrained by local receptive fields, which makes modeling long-range dependencies more difficult without substantially increasing network depth or complexity \cite{rs13122275}. This limitation becomes particularly relevant in HSI, where nonlocal interactions across spatial regions and spectral patterns may carry important semantic information. Recent efforts have therefore explored more expressive architectures for capturing such broader contextual relationships, including efficient spectral-spatial designs and knowledge-distillation-based formulations \cite{11119702, 11222092}.

Transformer architectures offer a natural alternative because self-attention explicitly models interactions between tokens across the entire input \cite{10155242}. This capability has motivated a growing number of Transformer-based and hybrid models for HSIC. Recent representative examples include MSST \cite{11352814}, S2CIFT \cite{11386932}, and S2CAT \cite{11359183}, which strengthen spectral-spatial interaction through tailored tokenization and fusion strategies. Other studies have investigated wavelet-assisted and differential attention formulations, such as WaveFormer \cite{10399798} and DiffFormer \cite{10955699}. In parallel, state-space and Mamba-inspired models have emerged as powerful alternatives for long-range dependency modeling, including SSAM \cite{11355499}, differential Mamba-style attention \cite{11037730}, DBMLLA \cite{11150494}, WDM \cite{11083631}, and morphological Mamba variants \cite{AHMAD2025129995}. Further developments, such as GraphMamba \cite{10746459}, spiking Mamba \cite{11392780}, PolicyMamba \cite{11090003}, byte latent Mamba with knowledge distillation \cite{11222092}, graph-tokenized Mamba \cite{11226902}, and domain-adaptive Mamba \cite{11367376}, continue to demonstrate the growing importance of sequence modeling paradigms in hyperspectral analysis.

Despite their architectural diversity, most of these methods still rely on scaled dot-product attention and closely related inner-product-based similarity measures \cite{11195861,9767615,11129658,Ahmad18072025}. In this formulation, the interaction between query and key tokens is governed by their inner product in the embedding space, meaning that both vector magnitude and orientation affect the final score. While this is computationally convenient and effective in many settings, it implicitly assumes a Euclidean feature geometry that may not be the most appropriate for hyperspectral data. In remote sensing practice, spectral similarity is often evaluated using angular measures because the direction of a spectral signature is frequently more informative than its absolute magnitude. This is the central idea behind metrics such as the spectral angle mapper (SAM), which are known to be more robust to illumination changes and sensor-induced intensity variations. Similar observations have also been discussed in recent hyperspectral modeling work, where angular consistency has been linked to more discriminative spectral representation learning.

This mismatch between conventional dot-product attention and the angular nature of hyperspectral signatures motivates a closer examination of attention scoring itself. In many existing models, the attention module is adopted as a standard component, while the geometry induced by the similarity function receives comparatively less focus. However, the score function determines how token relationships are measured, amplified, and propagated through the network. For hyperspectral data, this choice may be especially consequential because spectral magnitude can vary due to acquisition conditions, atmospheric effects, and local illumination, even when the underlying material identity remains unchanged. A similarity function that is less sensitive to such magnitude variations could therefore offer a more suitable inductive bias for HSIC.

Another practical challenge in hyperspectral learning is the limited availability of labeled samples \cite{11421023,ahmad2024multihead}. Pixel-level annotation of HSIs is expensive, labor-intensive, and often requires expert knowledge, which results in training sets where only a small fraction of pixels are labeled \cite{11370490,10604879,11414160,11105087,11424414}. Under such label-scarce conditions, highly expressive models with large parameter counts can overfit easily and may struggle to generalize across scenes or sampling protocols \cite{9903062,10685113}. This challenge is particularly relevant for attention-based architectures, whose performance depends not only on backbone design but also on how effectively token relationships are encoded when supervision is extremely limited \cite{1011453465055,ahmad20263dfourierbasedglobalfeature}. Consequently, there is strong motivation to design attention mechanisms that are robust, lightweight, and better aligned with hyperspectral feature geometry \cite{11146867}.

To address this issue, this work revisits attention scoring from a geometric perspective and investigate the role of similarity functions in hyperspectral Transformers. As illustrated in Fig.~\ref{fig:geometry_attention}, we introduce a cosine-normalized attention formulation in which query and key embeddings are first projected onto a unit hypersphere and their similarity is then computed using squared cosine similarity. By normalizing the embeddings, the resulting score emphasizes angular agreement while suppressing the effect of feature magnitude. In this way, the attention mechanism becomes more consistent with the geometry of hyperspectral signatures, where angular relations often better reflect material similarity than raw Euclidean interactions.

The proposed attention mechanism is integrated into a spatial-spectral Transformer architecture and evaluated under extremely limited supervision. Beyond proposing a simple architectural modification, this work also provides a controlled analysis of multiple attention score formulations within the same backbone, allowing the effect of similarity design to be isolated more clearly than in comparisons across entirely different models. Experimental results on three benchmark datasets show that cosine-based scoring consistently achieves higher performance and outperforms several recent Transformer- and Mamba-based models while retaining a lightweight design. Additional analyses, including stability evaluation, patch-size sensitivity, and robustness to spectral noise, further confirm the effectiveness of the proposed formulation and provide deeper insight into why geometry-aware scoring is beneficial for HSIC.

Overall, this study highlights that the similarity function used inside attention is not merely a minor implementation detail, but a meaningful inductive bias that shapes hyperspectral representation learning. Aligning attention scoring with the angular structure of hyperspectral signatures offers a simple, interpretable, and effective alternative to conventional dot-product attention, especially in low-label settings where robust generalization is essential.

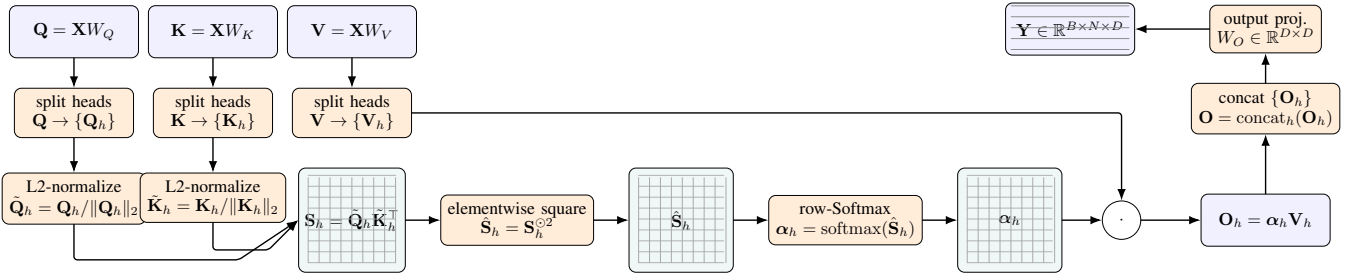
\begin{figure*}[!t]
\centering
\resizebox{0.99\textwidth}{!}{
\begin{tikzpicture}[
  >=Latex, line cap=round,
  font=\normalsize,
  mat/.style     ={draw,rounded corners,fill=blue!6,minimum width=2.6cm,minimum height=1.05cm},
  bigmat/.style  ={draw,rounded corners,fill=teal!6,minimum width=2.6cm,minimum height=2.1cm},
  gridbox/.style ={draw=gray!70, fill=gray!10, line width=0.3pt},
  sq/.style      ={draw=gray!70, fill=gray!25, minimum size=3.6mm, inner sep=0pt},
  op/.style      ={draw,rounded corners,fill=orange!15,inner sep=4pt},
  sum/.style     ={circle,draw,minimum size=6mm,inner sep=6pt},
  lab/.style     ={inner sep=1pt}
]
\node[mat,xshift=-28mm] (Qmat)
    {$\mathbf{Q} = \mathbf{X}W_Q$};
\node[mat] (Kmat)
    {$\mathbf{K} = \mathbf{X}W_K$};
\node[mat, xshift=28mm] (Vmat)
    {$\mathbf{V} = \mathbf{X}W_V$};

\node[op, below=6mm of Qmat, align=center, minimum width=2.4cm] (splitQ)
  {split heads \\[-1pt] $\mathbf{Q}\to\{\mathbf{Q}_h\}$};
\node[op, below=6mm of Kmat, align=center, minimum width=2.4cm] (splitK)
  {split heads \\[-1pt] $\mathbf{K}\to\{\mathbf{K}_h\}$};
\node[op, below=6mm of Vmat, align=center, minimum width=2.4cm] (splitV)
  {split heads \\[-1pt] $\mathbf{V}\to\{\mathbf{V}_h\}$};
\draw[-{Latex},thick] (Qmat.south) -- (splitQ.north);
\draw[-{Latex},thick] (Kmat.south) -- (splitK.north);
\draw[-{Latex},thick] (Vmat.south) -- (splitV.north);

\node[op, below=7mm of splitQ, align=center, minimum width=2.6cm] (normQ)
  {L2-normalize \\ $\tilde{\mathbf{Q}}_h=\mathbf{Q}_h/\|\mathbf{Q}_h\|_2$};
\node[op, below=7mm of splitK, align=center, minimum width=2.6cm] (normK)
  {L2-normalize \\[-1pt] $\tilde{\mathbf{K}}_h=\mathbf{K}_h/\|\mathbf{K}_h\|_2$};
\draw[-{Latex},thick] (splitQ.south) -- (normQ.north);
\draw[-{Latex},thick] (splitK.south) -- (normK.north);

\node[bigmat, right=10mm of normK, below=6mm of splitV, minimum width=2.1cm, minimum height=2.1cm]
    (Smat)
    {$\mathbf{S}_h=\tilde{\mathbf{Q}}_h\tilde{\mathbf{K}}_h^{\top}$};
\foreach \k in {0,...,8}{
  \draw[gray!70,opacity=0.85]
    ($(Smat.south west)+(0.20,0.20+0.22*\k)$) --
    ($(Smat.south east)+(-0.20,0.20+0.22*\k)$);
  \draw[gray!70,opacity=0.95]
    ($(Smat.south west)+(0.20+0.22*\k,0.20)$) --
    ($(Smat.north west)+(0.20+0.22*\k,-0.20)$);
}
\coordinate (SinQ) at ($(Smat.west)+(-1.2,  -0.8)$);
\coordinate (SinK) at ($(Smat.west)+(-0.6, -0.6)$);
\draw[-{Latex},thick,rounded corners=2pt]
  (normQ.south) |- (SinQ) -- (Smat.west);
\draw[-{Latex},thick,rounded corners=2pt]
  (normK.south) |- (SinK) -- (Smat.west);

\node[op, right=7mm of Smat, align=center, minimum width=1.8cm] (square)
  {elementwise square \\[-1pt] $\hat{\mathbf{S}}_h=\mathbf{S}_h^{\odot 2}$};
\draw[-{Latex},thick] (Smat.east) -- (square.west);
\node[bigmat, right=7mm of square, minimum width=2.1cm, minimum height=2.1cm]
    (Shat)
    {$\hat{\mathbf{S}}_h$};
\foreach \k in {0,...,8}{
  \draw[gray!70,opacity=0.85]
    ($(Shat.south west)+(0.20,0.20+0.22*\k)$) --
    ($(Shat.south east)+(-0.20,0.20+0.22*\k)$);
  \draw[gray!70,opacity=0.95]
    ($(Shat.south west)+(0.20+0.22*\k,0.20)$) --
    ($(Shat.north west)+(0.20+0.22*\k,-0.20)$);
}
\draw[-{Latex},thick] (square.east) -- (Shat.west);
\node[op, right=7mm of Shat, align=center, minimum width=1.8cm] (softmax)
  {row-Softmax \\[-1pt] $\boldsymbol{\alpha}_h=\mathrm{softmax}(\hat{\mathbf{S}}_h)$};
\draw[-{Latex},thick] (Shat.east) -- (softmax.west);

\node[bigmat, right=7mm of softmax, minimum width=2.1cm, minimum height=2.1cm] (Alpha) {$\boldsymbol{\alpha}_h$};
\foreach \k in {0,...,8}{
  \draw[gray!70,opacity=0.85]
    ($(Alpha.south west)+(0.20,0.20+0.22*\k)$) --
    ($(Alpha.south east)+(-0.20,0.20+0.22*\k)$);
  \draw[gray!70,opacity=0.95]
    ($(Alpha.south west)+(0.20+0.22*\k,0.20)$) --
    ($(Alpha.north west)+(0.20+0.22*\k,-0.20)$);
}
\draw[-{Latex},thick] (softmax.east) -- (Alpha.west);
\node[sum, right=8mm of Alpha] (dotAtt) {$\cdot$};
\draw[-{Latex},thick] (Alpha.east) -- (dotAtt.west);
\draw[-{Latex},thick,rounded corners=2pt]
  (splitV.east) -| (dotAtt.north);
\node[mat, right = 12mm of dotAtt] (Ohmat) {$\mathbf{O}_h=\boldsymbol{\alpha}_h\mathbf{V}_h$};
\draw[-{Latex},thick] (dotAtt.east) -- (Ohmat.west);

\node[op, above=12mm of Ohmat, align=center, minimum width=2.2cm] (merge)
  {concat $\{\mathbf{O}_h\}$ \\[-1pt] $\mathbf{O}=\mathrm{concat}_h(\mathbf{O}_h)$};
\draw[-{Latex},thick] (Ohmat.north) -- (merge.south);
\node[op, above=6mm of merge, align=center, minimum width=2.2cm] (Wout)
  {output proj. \\[-1pt] $W_O\in\mathbb{R}^{D\times D}$};
\draw[-{Latex},thick] (merge.north) -- (Wout.south);

\node[mat, left=15mm of Wout] (Ymat) {$\mathbf{Y}\in\mathbb{R}^{B\times N\times D}$};
\foreach \k in {0,...,4}{
  \draw[gray!90]
    ($(Ymat.south west)+(0.12,0.12+0.22*\k)$) --
    ($(Ymat.south east)+(-0.12,0.12+0.22*\k)$);
}
\draw[-{Latex},thick] (Wout.west) -- (Ymat.east);
\end{tikzpicture}
}
\caption{The HSI cube is partitioned into 3D patches, embedded into a token matrix, projected into query, key, and value representations, normalized, and scored using a cosine-based overlap function.}
\label{fig:qi_mha}
\end{figure*}

\begin{algorithm}[!t]
\caption{Cosine-Normalized Attention}
\label{alg:qimha}
\begin{algorithmic}[1]
\REQUIRE Token matrix $\mathbf{T}_{\ell}\in\mathbb{R}^{N\times D}$, score variant $v$
\FOR{each head $h=1,\dots,H$}
    \STATE $\mathbf{Q}_h,\mathbf{K}_h,\mathbf{V}_h \leftarrow$ head projections
    \IF{$v \in \{\text{cosine}, \text{cosine}^2, \text{cross-cosine}, \text{cross-cosine}^2\}$}
        \STATE $\tilde{\mathbf{Q}}_{h} = \mathbf{Q}_{h}/\|\mathbf{Q}_{h}\|_{2}$; ~~  $\tilde{\mathbf{K}}_{h} = \mathbf{K}_{h}/\|\mathbf{K}_{h}\|_{2}$
    \ENDIF
    \IF{$v = \text{cosine}^2$}
        \STATE $\mathbf{S}_{h} = \tilde{\mathbf{Q}}_{h}\tilde{\mathbf{K}}_{h}^{\top}$; ~~ $\mathbf{A}_{h} = \mathbf{S}_{h}^{\odot 2}$
    \ELSIF{$v = \text{cosine}$}
        \STATE $\mathbf{A}_{h} = \tilde{\mathbf{Q}}_{h}\tilde{\mathbf{K}}_{h}^{\top}$
    \ELSIF{$v = \text{scaled dot-product}$}
        \STATE $\mathbf{A}_{h} = (\mathbf{Q}_{h}\mathbf{K}_{h}^{\top})/\sqrt{d_h}$
    \ELSIF{$v = \text{dot-product}$}
        \STATE $\mathbf{A}_{h} = \mathbf{Q}_{h}\mathbf{K}_{h}^{\top}$
    \ELSIF{$v = \text{additive}$}
        \STATE $a_{ij,h} = \mathbf{w}_{a,h}^{\top}\tanh(\mathbf{W}_{q,h}\mathbf{q}_{i,h}+\mathbf{W}_{k,h}\mathbf{k}_{j,h}+\mathbf{b}_{a,h})$; ~~ $\mathbf{A}_{h} \leftarrow \{a_{ij,h}\}$
    \ELSIF{$v = \text{cross-cosine}^2$}
        \STATE $\mathbf{S}_{h} = \tilde{\mathbf{Q}}_{h}\tilde{\mathbf{K}}_{h}^{\top}$
        \STATE $\mathbf{A}_{h} = \mathbf{S}_{h}^{\odot 2}$
    \ELSIF{$v = \text{cross-cosine}$}
        \STATE $\mathbf{A}_{h} = \tilde{\mathbf{Q}}_{h}\tilde{\mathbf{K}}_{h}^{\top}$
    \ELSIF{$v = \text{cross-scaled dot-product}$}
        \STATE $\mathbf{A}_{h} = (\mathbf{Q}_{h}\mathbf{K}_{h}^{\top})/\sqrt{d_h}$
    \ELSIF{$v = \text{cross-additive}$}
        \STATE $a_{ij,h} = \mathbf{w}_{a,h}^{\top}\tanh(\mathbf{W}_{q,h}\mathbf{q}_{i,h}+\mathbf{W}_{k,h}\mathbf{k}_{j,h}+\mathbf{b}_{a,h})$; ~~ $\mathbf{A}_{h} \leftarrow \{a_{ij,h}\}$
    \ENDIF
    \STATE $\boldsymbol{\alpha}_{h} = \mathrm{softmax}(\mathbf{A}_{h})$
    \STATE $\mathbf{O}_{h} = \boldsymbol{\alpha}_{h}\mathbf{V}_{h}$
\ENDFOR
\STATE Concatenate head outputs:
$\mathbf{O}=[\mathbf{O}_1\|\dots\|\mathbf{O}_H]$
\STATE Output projection:
$\mathbf{Y}_{\ell}=\mathbf{O}\mathbf{W}_O$
\RETURN $\mathbf{Y}_{\ell}$
\end{algorithmic}
\end{algorithm}

\section{Proposed Methodology}

Let the HSI be denoted by $\mathbf{I}\in\mathbb{R}^{H\times W\times C}$, where $H$ and $W$ represent the spatial dimensions and $C$ denotes the number of spectral bands. To jointly exploit spatial context and spectral information, we adopt a patch-based spatial--spectral Transformer framework. Overlapping patches of size $P\times P\times C$ are extracted from the HSI and processed independently by the proposed model. The overall architecture is illustrated in Fig.~\ref{fig:qi_mha}, and  Algorithm~\ref{alg:qimha}.

\subsection{Patch Tokenization and Spectral Embedding}

Let $\mathbf{X}\in\mathbb{R}^{P\times P\times C}$ denote an input hyperspectral patch. For each spatial location $(i,j)$, the spectral measurement is $\mathbf{x}_{i,j}\in\mathbb{R}^{C}$. Each spectral vector is projected into a $D$-dimensional embedding space using a learnable linear transformation $\mathbf{W}_s\in\mathbb{R}^{C\times D}$ as $\mathbf{z}_{i,j} = \mathbf{W}_s^{\top}\mathbf{x}_{i,j}$. Stacking the embeddings over the spatial patch yields $\mathbf{Z}\in\mathbb{R}^{P\times P\times D}$, which is reshaped into a sequence of $N=P^2$ tokens $\mathbf{T} = \mathrm{reshape}(\mathbf{Z}) \in \mathbb{R}^{N\times D}$. To retain spatial structure, a learnable positional embedding $\mathbf{E}\in\mathbb{R}^{N\times D}$ is added $\mathbf{T}_0 = \mathbf{T} + \mathbf{E}$. 

\subsection{Cosine-Normalized Multi-Head Attention}

Given the token matrix $\mathbf{T}_\ell\in\mathbb{R}^{N\times D}$ at encoder layer $\ell$, query, key, and value matrices are computed as $\mathbf{Q}=\mathbf{T}_\ell\mathbf{W}_Q$, $\mathbf{K}=\mathbf{T}_\ell\mathbf{W}_K$, and $\mathbf{V}=\mathbf{T}_\ell\mathbf{W}_V$. These matrices are partitioned into $H$ attention heads, where each head has dimension $d_h=D/H$. Fig.~\ref{fig:geometry_attention} illustrates the proposed normalization at the level of a single query--key pair. In the actual attention module, this operation is applied row-wise to the query and key matrices of each head. Thus, for the $i$th query vector and $j$th key vector in head $h$, we compute
\begin{equation}
\tilde{\mathbf{q}}_{i,h} = \frac{\mathbf{q}_{i,h}}{\|\mathbf{q}_{i,h}\|_2},\qquad
\tilde{\mathbf{k}}_{j,h} = \frac{\mathbf{k}_{j,h}}{\|\mathbf{k}_{j,h}\|_2}
\end{equation}
Stacking these normalized vectors yields $\tilde{\mathbf{Q}}_h$ and $\tilde{\mathbf{K}}_h$, so that attention scores depend primarily on angular relationships between spectral features. More generally, the attention score matrix for head $h$ can be written as
\begin{equation}
\mathbf{A}_{h}=f_{\mathrm{att}}(\mathbf{Q}_{h},\mathbf{K}_{h})
\end{equation}
where $f_{\mathrm{att}}(\cdot)$ denotes the selected score function. The attention weights are then obtained through row-wise softmax $\boldsymbol{\alpha}_{h} = \mathrm{softmax}(\mathbf{A}_{h}), ~ \boldsymbol{\alpha}_{h}\in\mathbb{R}^{N\times N}$. 

\paragraph{\textbf{Self-attention variants}} For the principal formulation, similarity is computed using cosine similarity $\mathbf{S}_h = \tilde{\mathbf{Q}}_h \tilde{\mathbf{K}}_h^{\top}$, and the score matrix is obtained by element-wise squaring:
\begin{equation}
\mathbf{A}_h = \mathbf{S}_h^{\odot 2}
\end{equation}
We also evaluate cosine similarity without squaring, dot-product attention $\mathbf{A}_{h} = \mathbf{Q}_{h}\mathbf{K}_{h}^{\top}$ and scaled dot-product attention
\begin{equation}
\mathbf{A}_{h} = \frac{\mathbf{Q}_{h}\mathbf{K}_{h}^{\top}}{\sqrt{d_h}}
\end{equation}
and additive attention
\begin{equation}
a_{ij,h} = \mathbf{w}_{a,h}^{\top}
\tanh\!\left(
\mathbf{W}_{q,h}\mathbf{q}_{i,h} +
\mathbf{W}_{k,h}\mathbf{k}_{j,h} +
\mathbf{b}_{a,h}
\right)
\end{equation}
where $\mathbf{q}_{i,h}$ and $\mathbf{k}_{j,h}$ denote the $i$th query and $j$th key vectors of head $h$, respectively.

\paragraph{\textbf{Cross-attention variants}} To examine whether normalized similarity remains beneficial when queries and keys originate from different token streams, cross-attention variants are also considered. Let $\mathbf{T}_{\ell}^{(q)}$ and $\mathbf{T}_{\ell}^{(k,v)}$ denote the query and key/value token streams. The corresponding projections are
\begin{equation}
\mathbf{Q}=\mathbf{T}_{\ell}^{(q)}\mathbf{W}_{Q},~~
\mathbf{K}=\mathbf{T}_{\ell}^{(k,v)}\mathbf{W}_{K},~~
\mathbf{V}=\mathbf{T}_{\ell}^{(k,v)}\mathbf{W}_{V}
\end{equation}
Under this setting, scaled dot-product, cosine, squared cosine, and additive attention are evaluated.

\paragraph{\textbf{Value aggregation and output projection}} Each head produces an output representation $\mathbf{O}_h = \boldsymbol{\alpha}_h \mathbf{V}_h \in\mathbb{R}^{N\times d_h}$. The head outputs are concatenated and projected back to the model dimension $\mathbf{Y}_{\ell} = \big[\mathbf{O}_{1}\,\|\,\dots\,\|\,\mathbf{O}_{H}\big]\mathbf{W}_{O}, ~\mathbf{W}_{O}\in\mathbb{R}^{D\times D}$.

Each encoder block consists of a cosine-normalized attention layer followed by a feed-forward network as $\mathbf{U}_\ell = \mathbf{T}_\ell + \mathrm{Dropout}\!\left(\mathrm{Attn}(\mathrm{LN}(\mathbf{T}_\ell))\right)$ and $\mathbf{T}_{\ell+1} = \mathbf{U}_\ell + \mathrm{Dropout}\!\left(\mathrm{MLP}(\mathrm{LN}(\mathbf{U}_\ell))\right)$. The final token representations are normalized and aggregated using global average pooling $\mathbf{g} = \mathrm{GAP}(\mathrm{LN}(\mathbf{T}_L))$. The pooled feature vector is then mapped to class probabilities $\hat{\mathbf{y}}=\mathrm{softmax}(\mathbf{W}_c^{\top}\mathbf{g}+\mathbf{b}_c)$. This lightweight classifier ensures that performance differences mainly reflect the effect of the attention formulation.

\section{Experimental Results and Discussion}

To evaluate the proposed cosine-normalized attention framework, experiments were conducted on three widely used HSI benchmarks: Salinas (SA), WHU\_Hi\_HongHu (HH), and QUH-Tangdaowan (TD). These datasets cover different scene complexities, class distributions, and spectral characteristics, providing a representative testbed for hyperspectral classification. Following a label-scarce setting, each dataset was divided into 1\% training, 1\% validation, and 98\% testing samples.

The proposed model was trained with patch size $WS=16$, batch size 128, and 50 epochs. The Transformer backbone used an embedding dimension of 64, depth 4, number of heads 4, and MLP dimension 128. Dropout was set to 0.1, with no additional dropout applied inside the attention score computation. Optimization was performed using AdamW with learning rate $3\times10^{-4}$, weight decay $2\times10^{-4}$, and gradient clipping with clipnorm $=1.0$. The classification objective was label-smoothed categorical cross-entropy with a smoothing factor of 0.05.

For comparison, the proposed model was evaluated against several recent HSIC methods, including MSST, S2CIFT, S2CAT, WaveFormer (WF), and Differential Transformer (DF). Furthermore, to reflect the latest advancements in state-space modeling, we have also added several recent Mamba-based methods, including SSAM, DBMLLA, WDM, MM, GraphMamba (GM), PolicyMamba (PM), and Byte Latent Mamba with Knowledge Distillation (KDM).

For the proposed framework, eleven attention score variants were evaluated within the same spatial-spectral Transformer backbone so that the effect of the scoring function could be examined under a controlled setting. For compact presentation, the evaluated variants are denoted as cosine-squared (CS$^2$), cosine similarity (CS), scaled dot-product (SDP), dot-product (DP), additive (Add), self-attention (SA), mixed self-attention with cosine-squared (MSA-CS$^2$), cross scaled dot-product (C-SDP), cross cosine-squared (C-CS$^2$), cross cosine (C-CS), and cross additive (C-Add).

\begin{table*}[!hbt]
    \centering
    \caption{Comparison of the proposed framework. (\gold{Highest}, \silver{Second}, and \bronze{Third highest}).}
    \resizebox{\textwidth}{!}{\begin{tabular}{c||ccccc||ccccccc||ccccccccccc} \hline \hline
        \multirow{2}{*}{\textbf{Metric}} & \multicolumn{5}{c||}{Transformers} & \multicolumn{7}{c||}{State Space Models} & \multicolumn{11}{c}{\textbf{Proposed}} \\ \cline{2-24}
        & MSST & S2CIFT & S2CAT & WF & DF & SSAM & DBMLLA & WDM & MM & GM & PM & KDM & CS$^2$ & CS & SDP & DP & Add & SA & MSA-CS$^2$ & C-SDP & C-CS$^2$ & C-CS & C-Add \\ \hline \hline 
        
        \multicolumn{24}{c}{SA Dataset} \\ \hline \hline
        $\kappa$ & 97.97 & 95.65 & 95.78 & 94.90 & 94.06 & 87.80 & 92.28 & 89.26 & 88.38 & 95.85 & 98.16 & 92.44 & 99.15 & 98.60 & 99.18 & 97.75 & 98.93 & \silver{99.24} & \bronze{99.23} & 98.76 & \gold{99.33} & 98.16 & 98.75 \\
        OA & 98.17 & 96.10 & 96.21 & 95.42 & 94.67 & 89.07 & 93.06 & 90.35 & 89.56 & 96.27 & 98.35 & 93.22 & 99.23 & 98.74 & 99.26 & 97.98 & 99.04 & \silver{99.32} & 99.31 & 98.88 & \gold{99.40} & 98.35 & 98.87 \\
        AA & 97.63 & 95.56 & 97.48 & 95.14 & 95.87 & 88.59 & 95.44 & 91.80 & 87.76 & 96.24 & 97.55 & 91.23 & \bronze{99.18} & 98.83 & 99.06 & 98.02 & 99.03 & \silver{99.20} & \gold{99.30} & 98.99 & 99.18 & 98.53 & 99.09 \\ \hline \hline

        \multicolumn{24}{c}{HH Dataset} \\ \hline \hline
        $\kappa$ & 95.48 & 89.18 & 88.31 & 88.32 & 83.70 & 83.40 & 79.56 & 79.23 & 77.48 & 93.70 & 65.50 & 84.35 & 96.94 & 97.02 & \gold{97.87} & 97.41 & 97.30 & \silver{97.68} & 97.21 & \bronze{97.63} & 96.29 & 97.22 & 97.09 \\
        OA & 96.42 & 91.43 & 90.78 & 90.76 & 87.11 & 86.94 & 84.01 & 83.66 & 82.16 & 95.01 & 73.56 & 87.61 & 97.58 & 97.64 & \gold{98.32} & 97.95 & 97.87 & \silver{98.16} & 97.80 & \bronze{98.12} & 97.06 & 97.80 & 97.69 \\
        AA & 93.19 & 78.77 & 79.30 & 80.48 & 69.60 & 67.74 & 53.42 & 58.30 & 57.95 & 88.92 & 32.51 & 65.35 & 93.05 & 92.38 & \silver{94.48} & 93.77 & 92.79 & \gold{94.86} & 93.74 & \bronze{94.21} & 88.62 & 92.72 & 93.74 \\ \hline \hline

        \multicolumn{24}{c}{TD Dataset} \\ \hline \hline
        $\kappa$ & 97.06 & 95.95 & 95.21 & 95.86 & 92.91 & 92.18 & 94.47 & 92.31 & 90.00 & 96.60 & 93.18 & 94.64 & 98.17 & \gold{98.68} & 98.30 & 98.19 & \bronze{98.51} & 98.32 & 98.31 & 98.21 & 98.38 & 98.41 & \silver{98.56} \\
        OA & 97.42 & 96.44 & 95.79 & 96.36 & 93.77 & 93.16 & 95.15 & 93.26 & 91.25 & 97.01 & 94.03 & 95.30 & 98.39 & \gold{98.84} & 98.51 & 98.41 & \bronze{98.69} & 98.52 & 98.52 & 98.43 & 98.58 & 98.60 & \silver{98.74} \\
        AA & 92.71 & 91.14 & 92.75 & 91.31 & 85.73 & 82.93 & 84.25 & 79.29 & 78.04 & 95.04 & 69.95 & 85.61 & 94.92 & \silver{97.23} & 95.88 & 93.90 & \bronze{96.51} & 93.82 & 95.80 & 95.41 & 95.51 & \gold{97.42} & 96.27 \\ \hline \hline
    \end{tabular}}
    \label{tab:placeholder}
\end{table*}

\begin{figure}[!hbt]
    \centering
    \subfloat[GT]{\includegraphics[width=0.13\columnwidth]{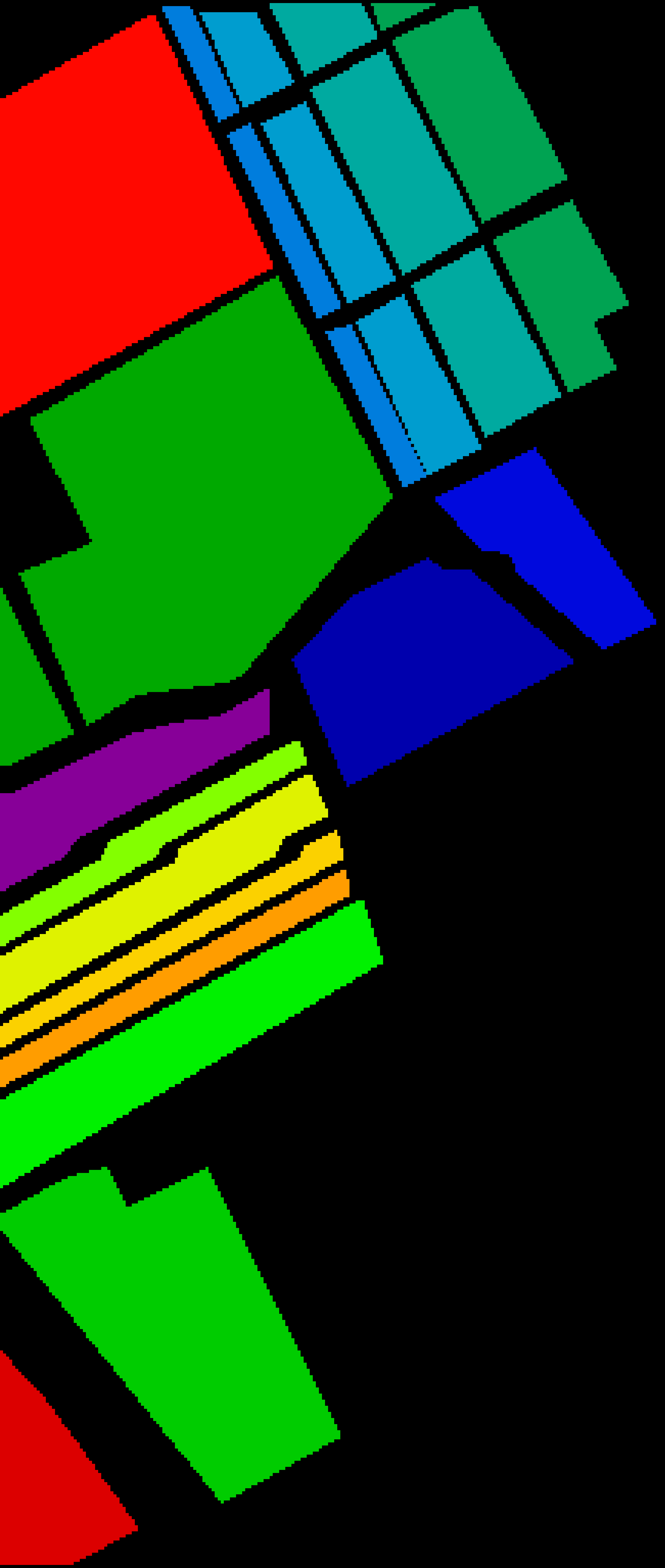}} \hspace{0.02cm}
    \subfloat[MSST]{\includegraphics[width=0.13\columnwidth]{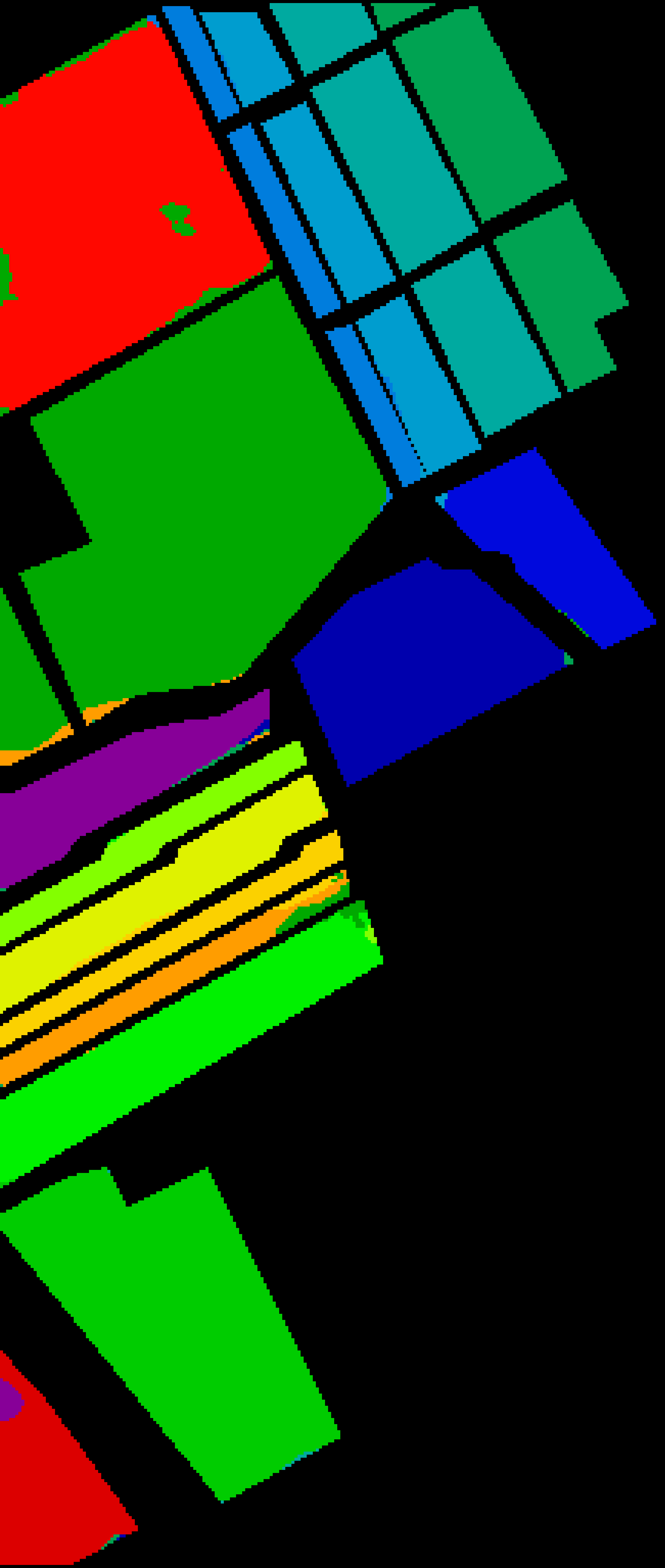}} \hspace{0.02cm}
    \subfloat[S2CIFT]{\includegraphics[width=0.13\columnwidth]{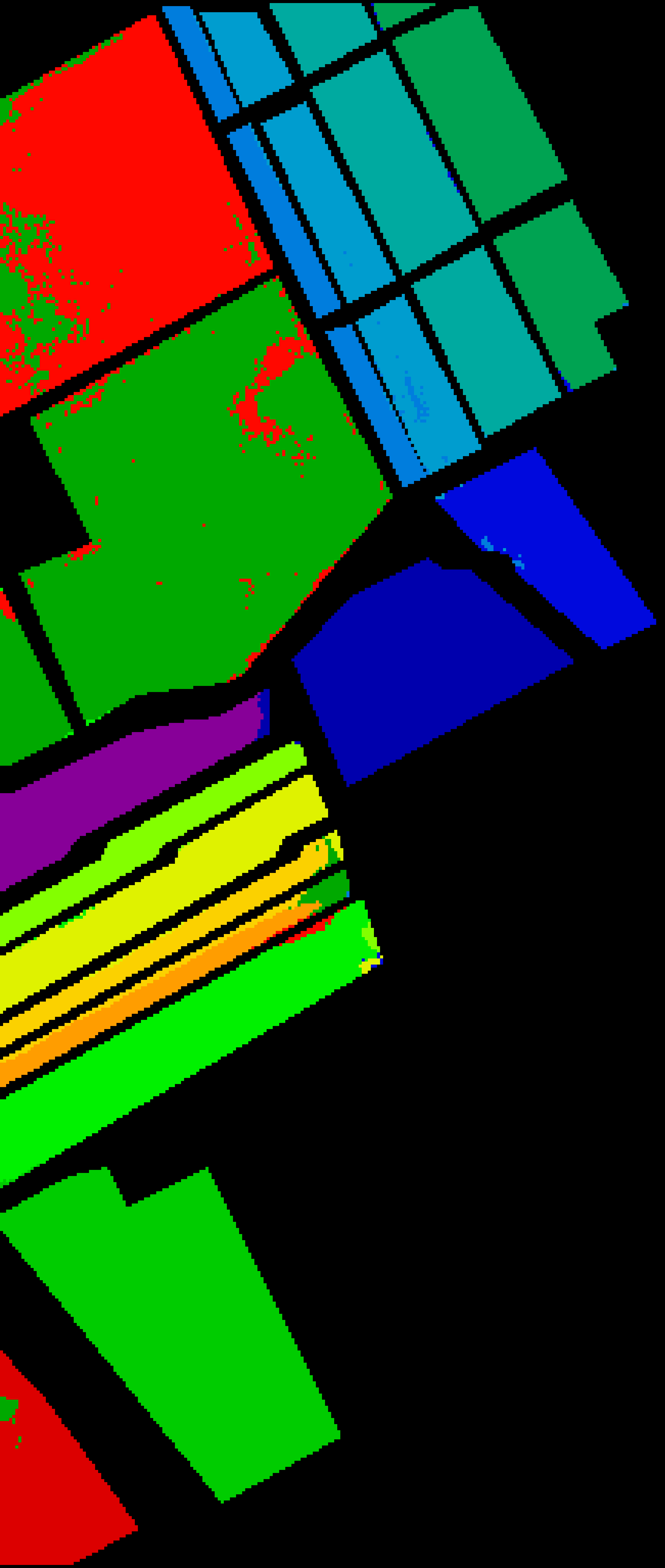}} \hspace{0.02cm}
    \subfloat[S2CAT]{\includegraphics[width=0.13\columnwidth]{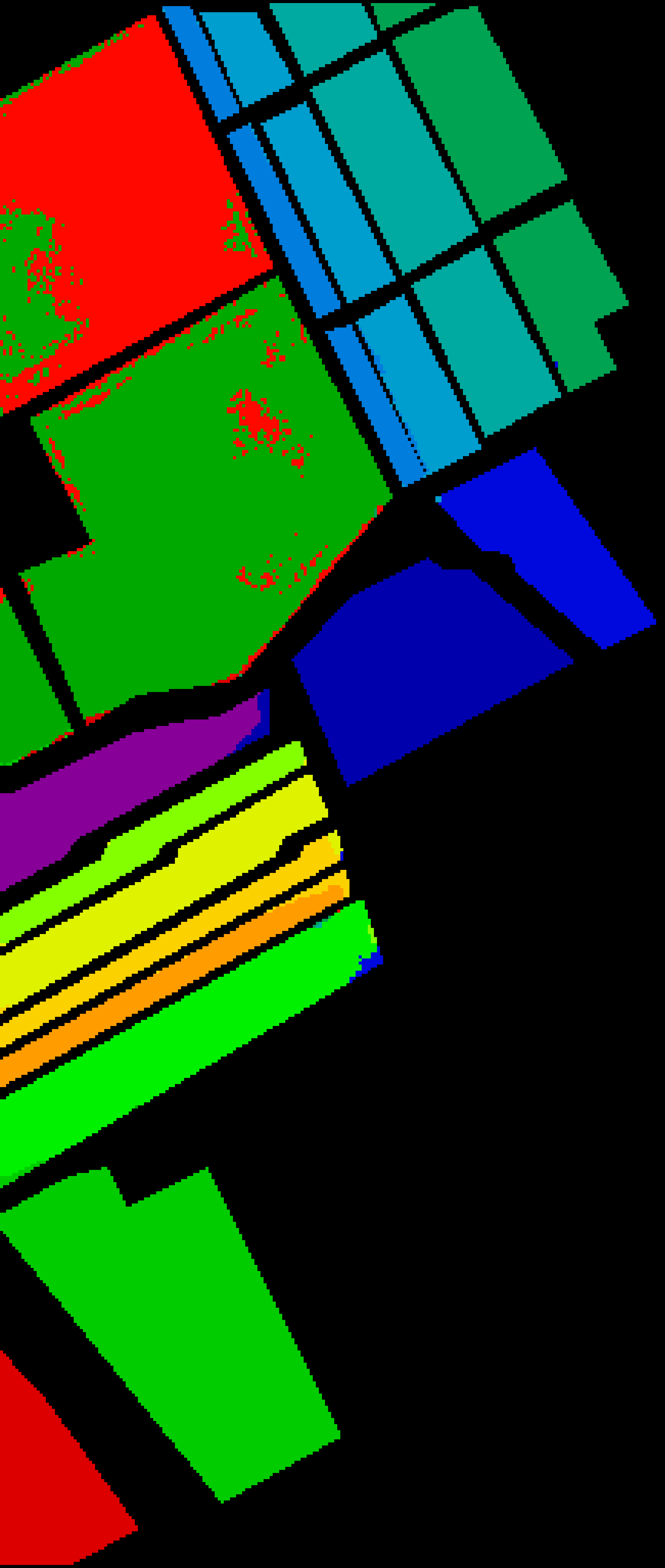}} \hspace{0.02cm}
    \subfloat[WF]{\includegraphics[width=0.13\columnwidth]{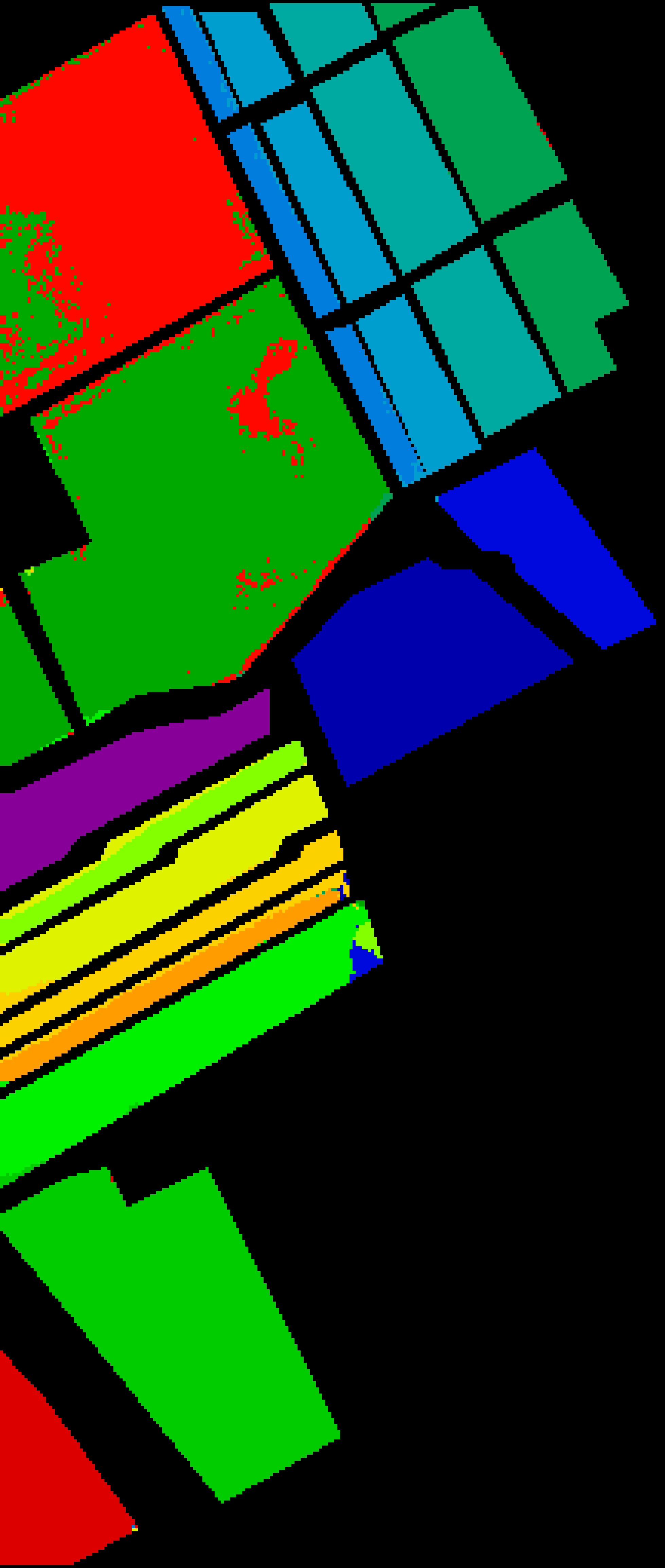}} \hspace{0.02cm}
    \subfloat[DF]{\includegraphics[width=0.13\columnwidth]{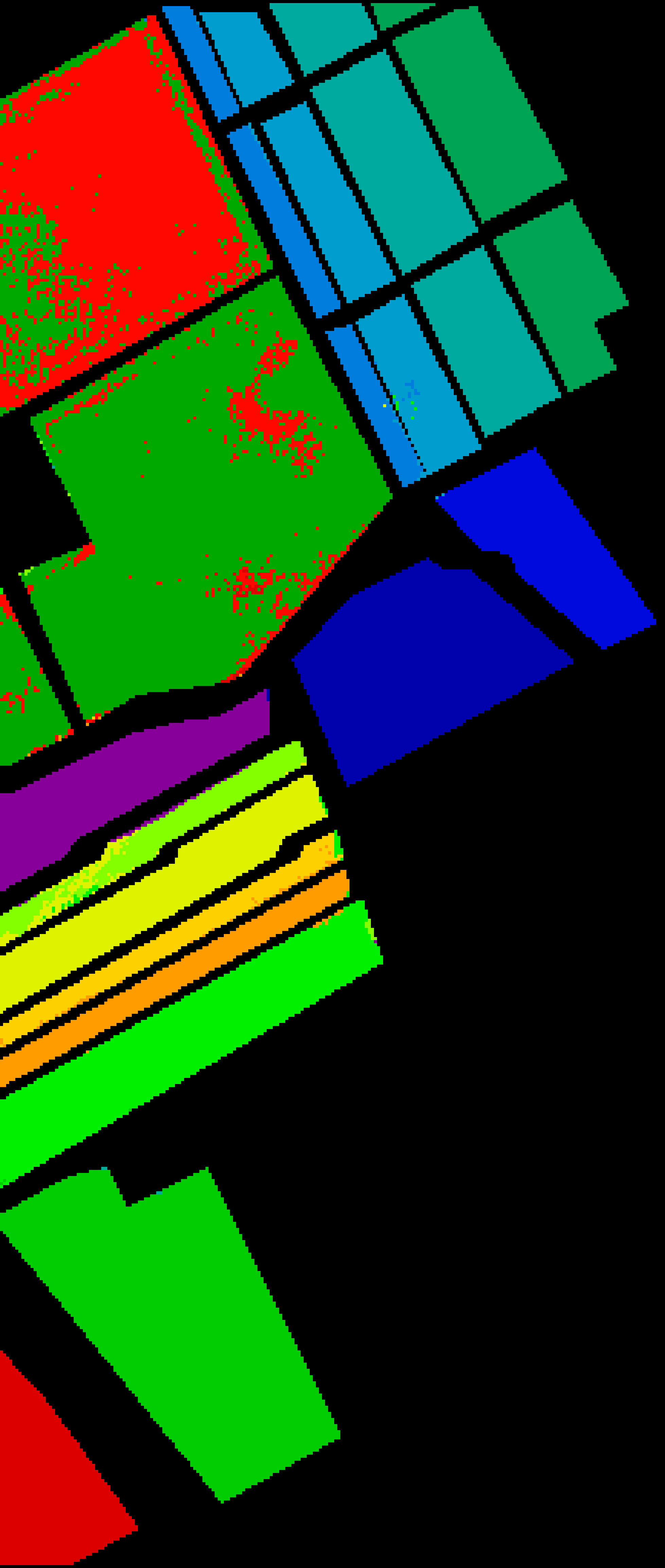}} \hspace{0.02cm}
    \subfloat[SSAM]{\includegraphics[width=0.13\columnwidth]{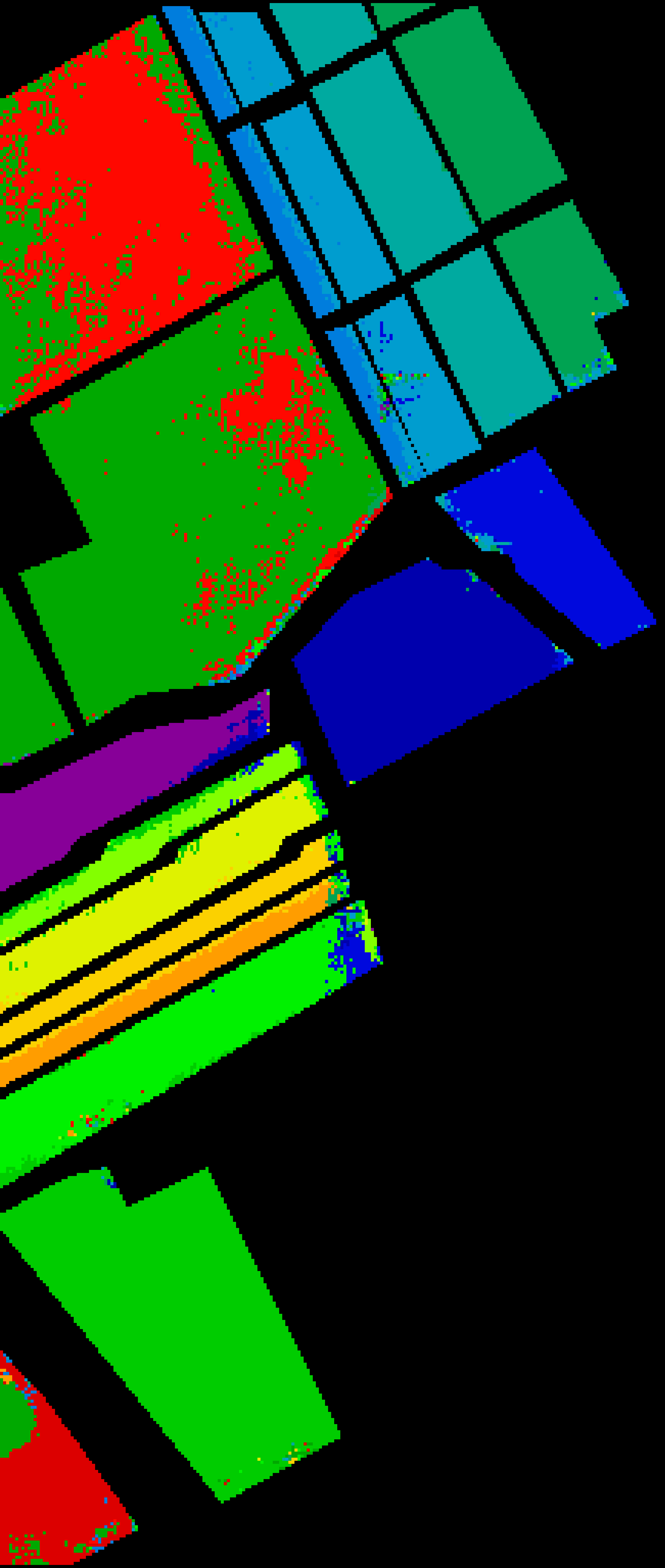}} \hspace{0.02cm}
    \subfloat[DBMLLA]{\includegraphics[width=0.13\columnwidth]{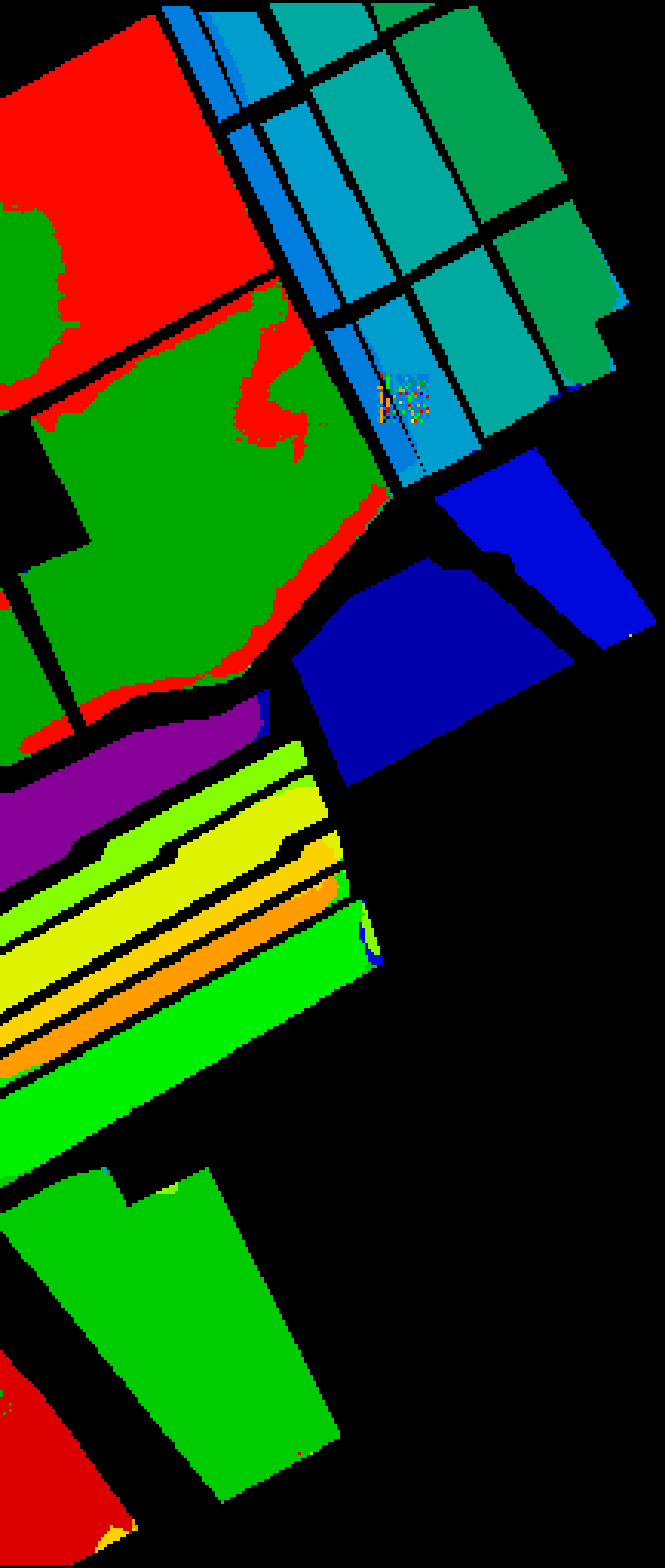}} \hspace{0.02cm}
    \subfloat[WDM]{\includegraphics[width=0.13\columnwidth]{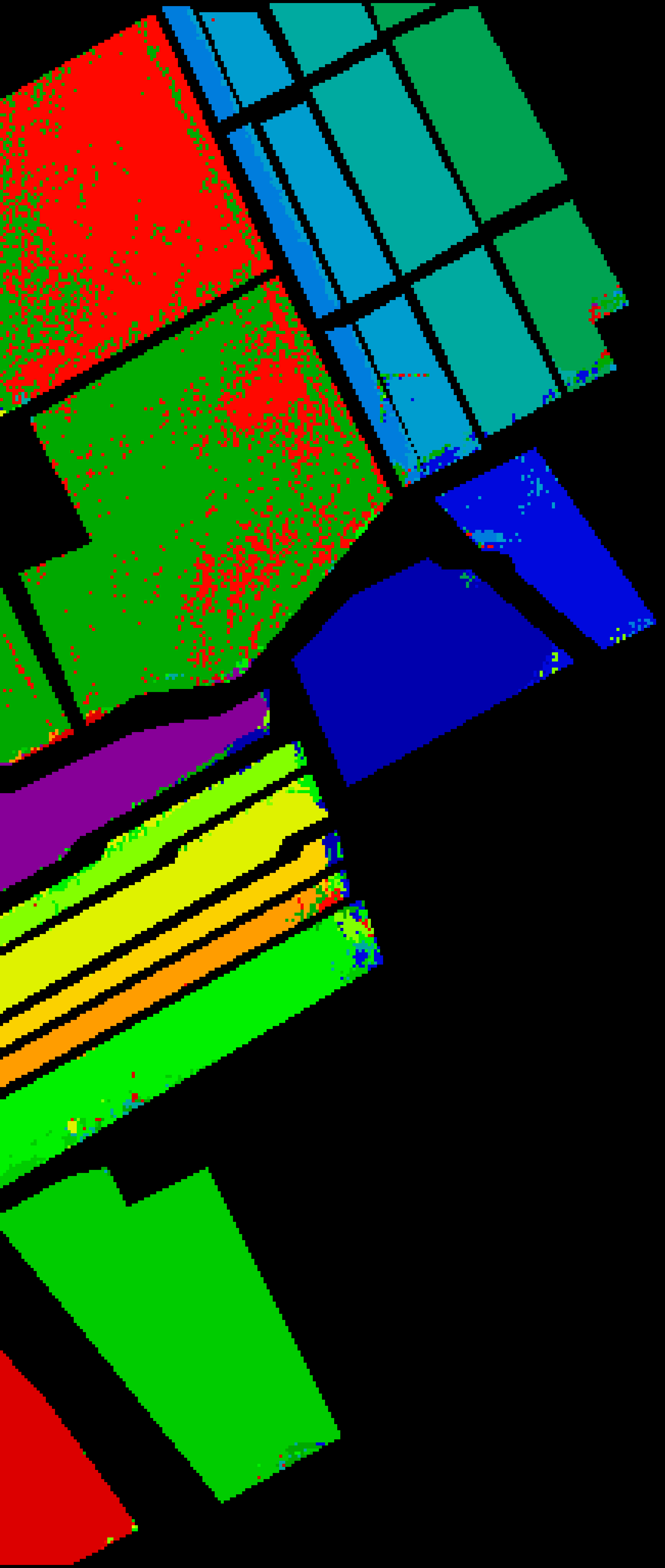}} \hspace{0.02cm}
    \subfloat[MM]{\includegraphics[width=0.13\columnwidth]{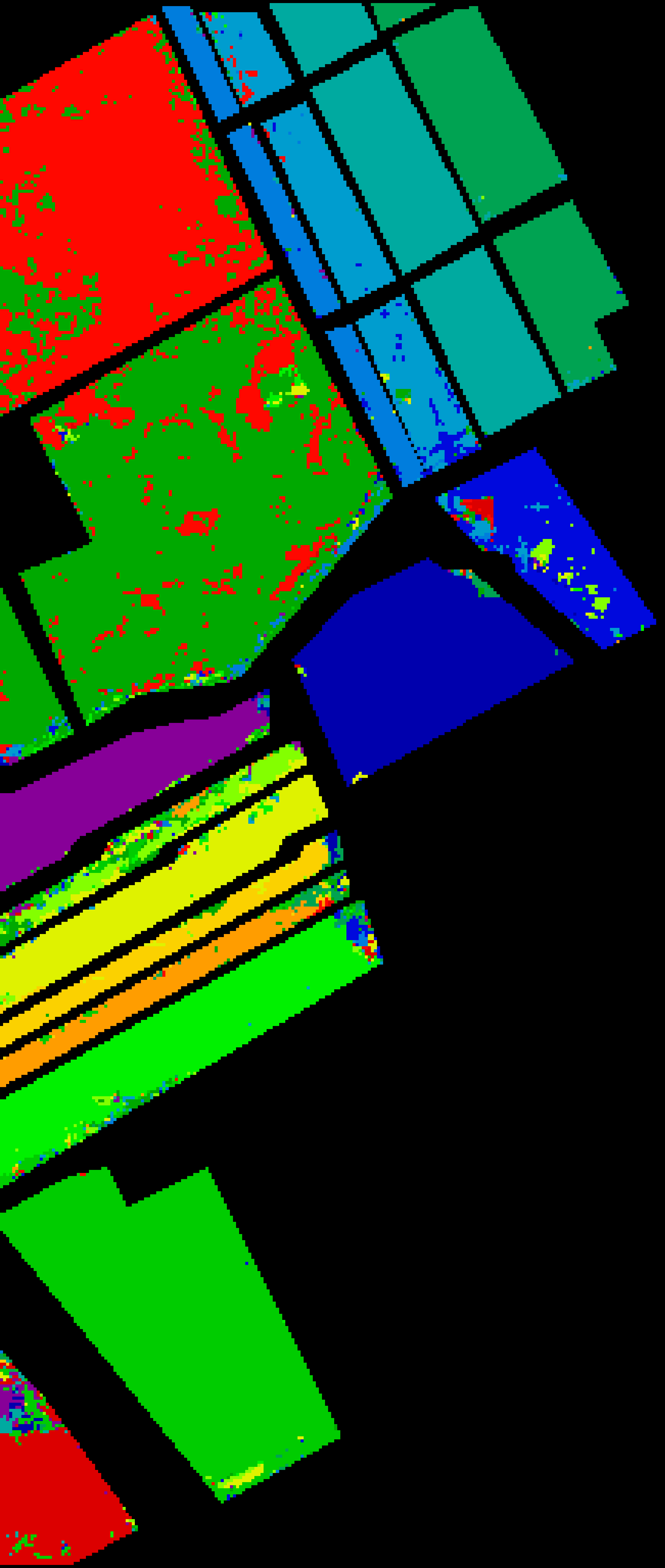}} \hspace{0.02cm}
    \subfloat[GM]{\includegraphics[width=0.13\columnwidth]{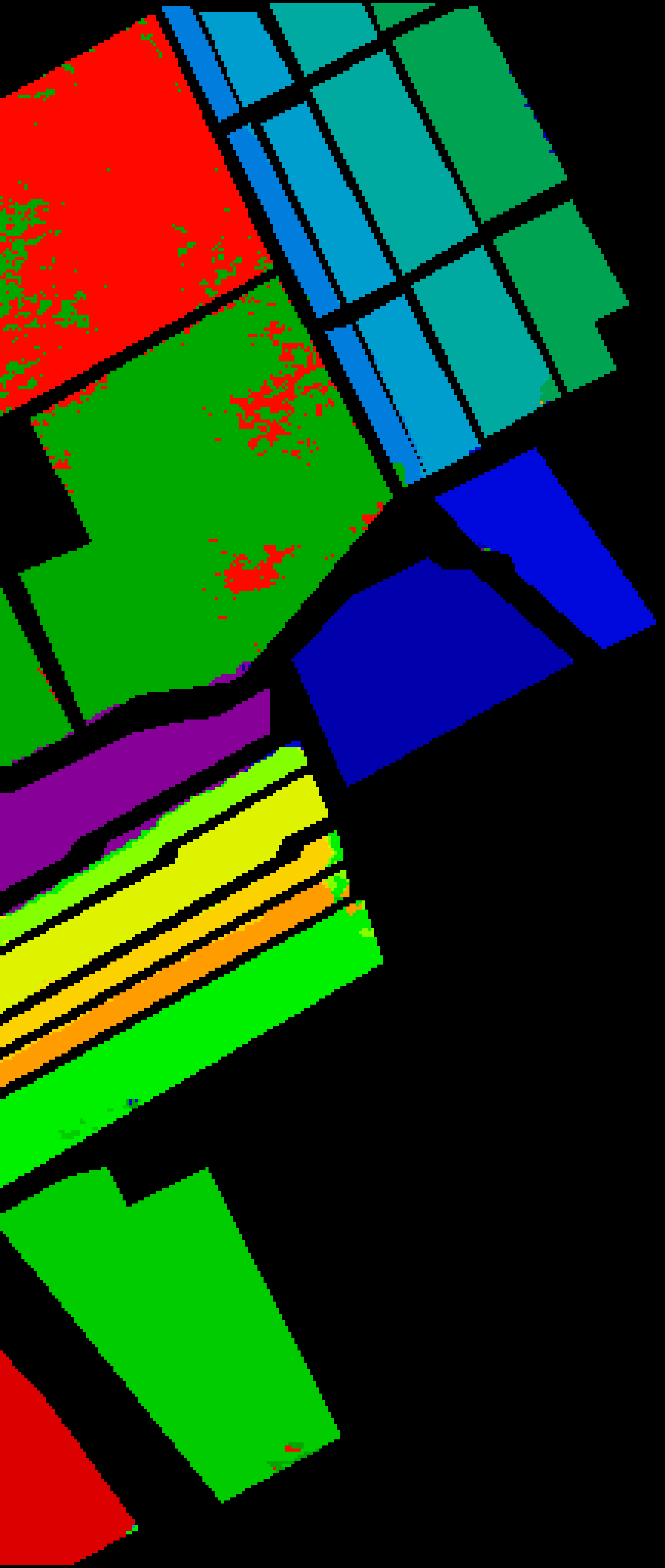}} \hspace{0.02cm}
    \subfloat[PM]{\includegraphics[width=0.13\columnwidth]{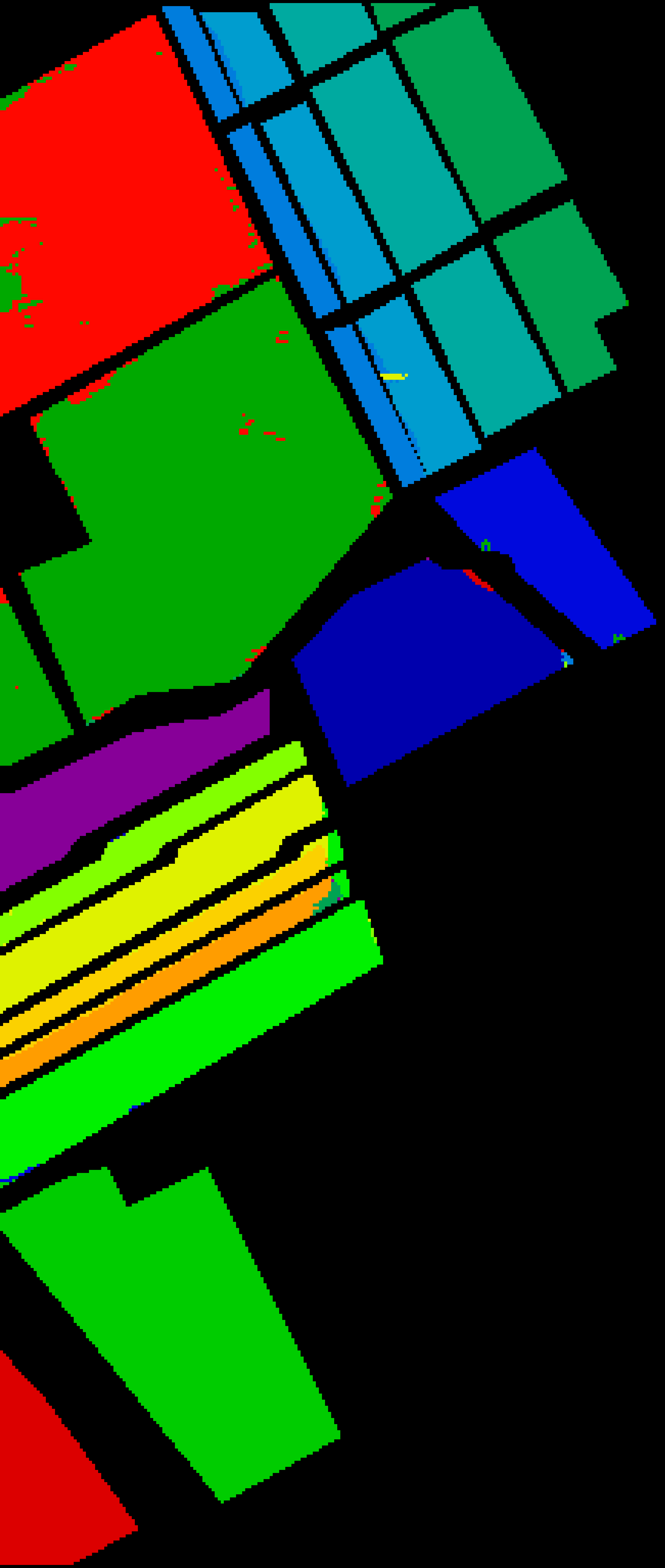}} \hspace{0.02cm}
    \subfloat[KDM]{\includegraphics[width=0.13\columnwidth]{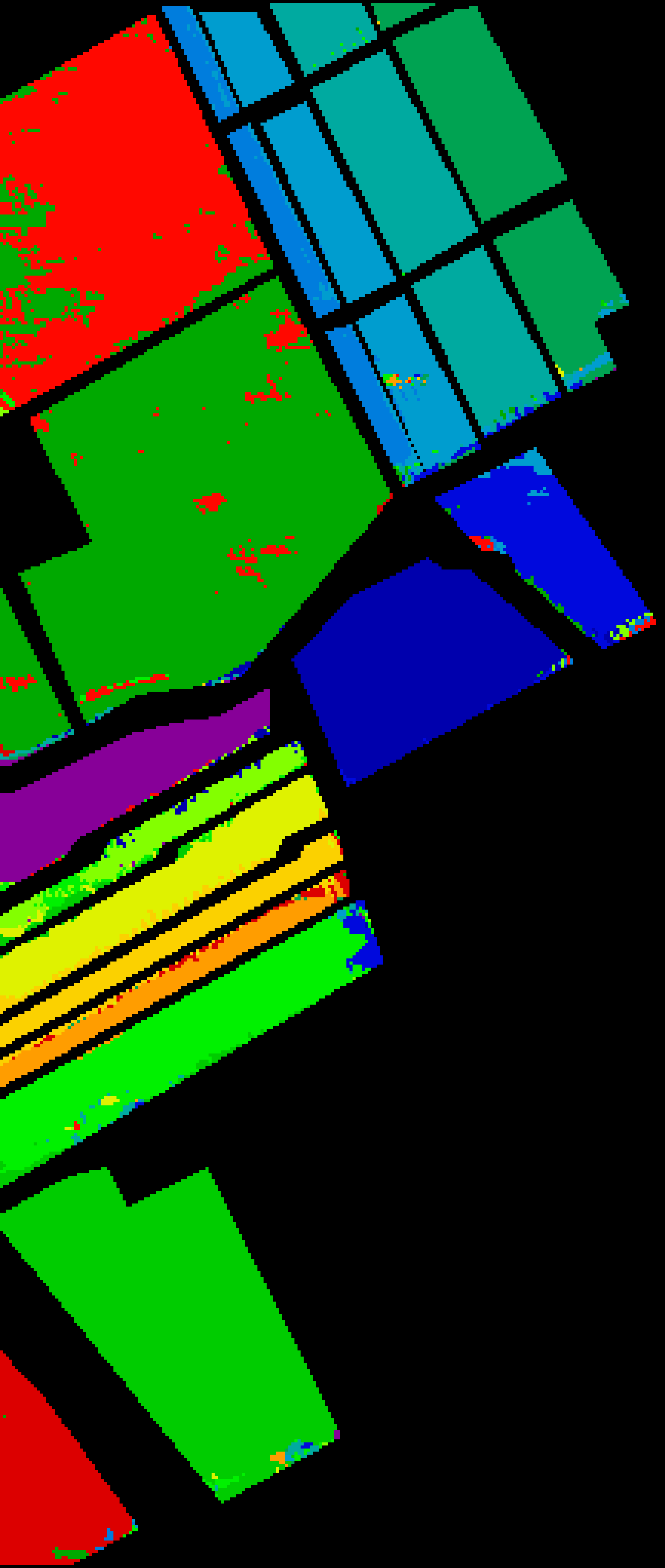}} \hspace{0.02cm}
    \subfloat[Pro.]{\includegraphics[width=0.13\columnwidth]{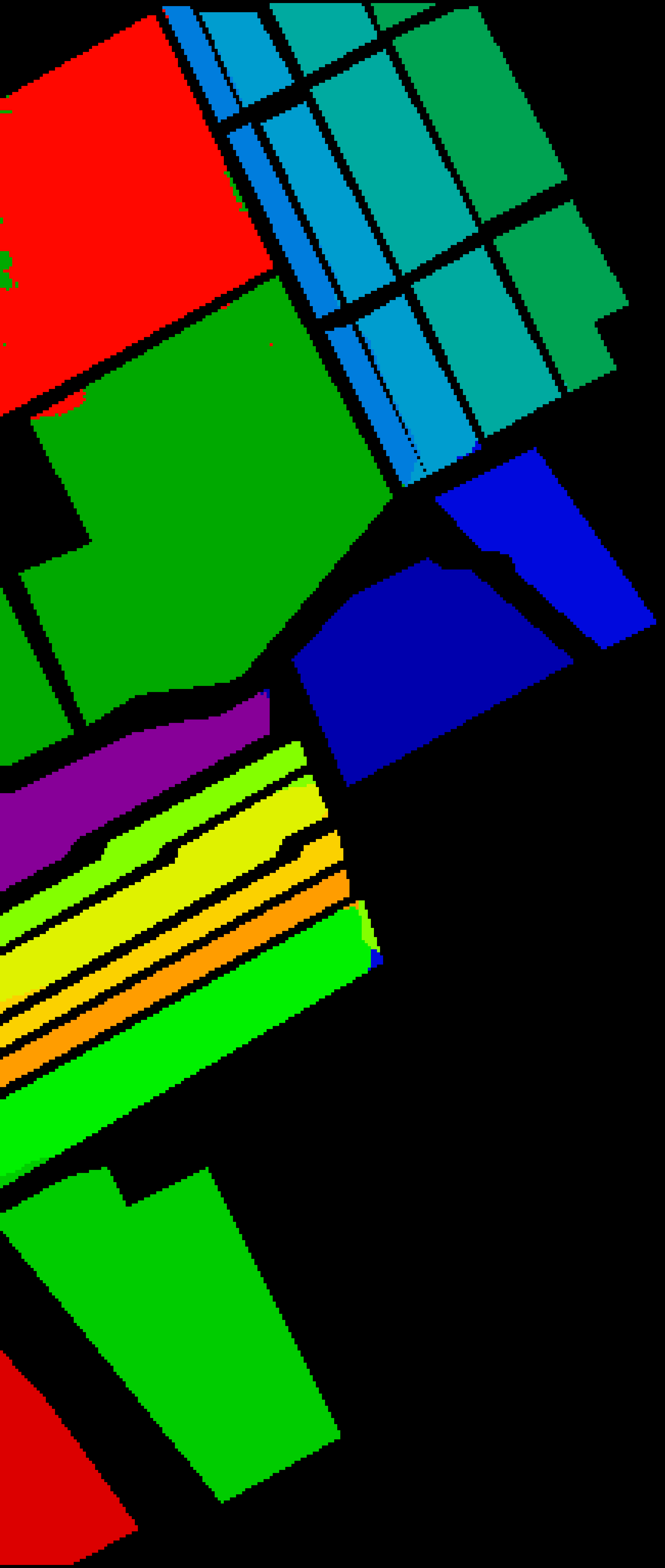}}
    \caption{Visual comparison of classification maps on SA.}
    \label{SAGT}
\end{figure}

\begin{figure}[!hbt]
    \centering
    \subfloat[GT]{\includegraphics[width=0.13\columnwidth]{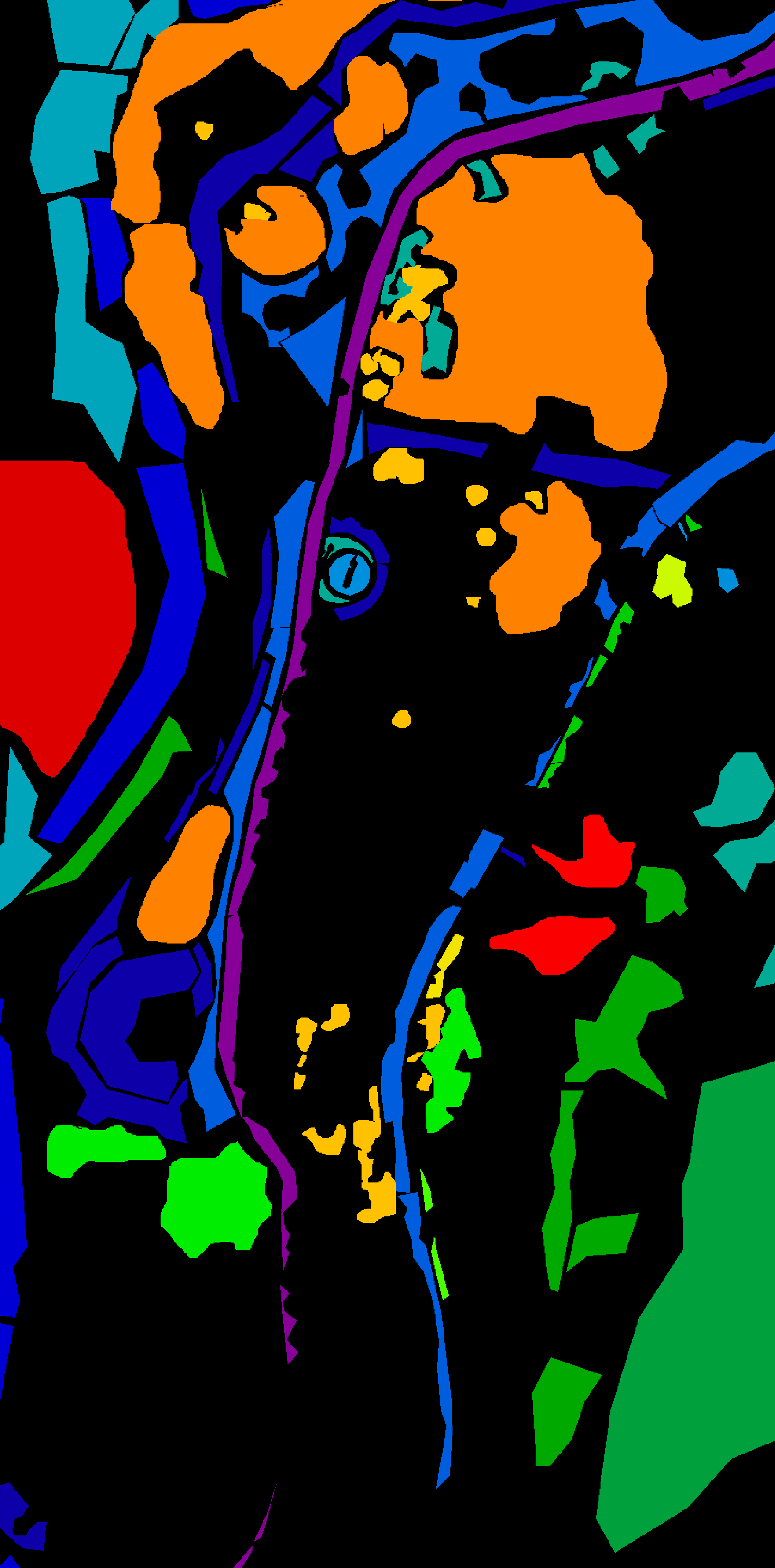}} \hspace{0.02cm}
    \subfloat[MSST]{\includegraphics[width=0.13\columnwidth]{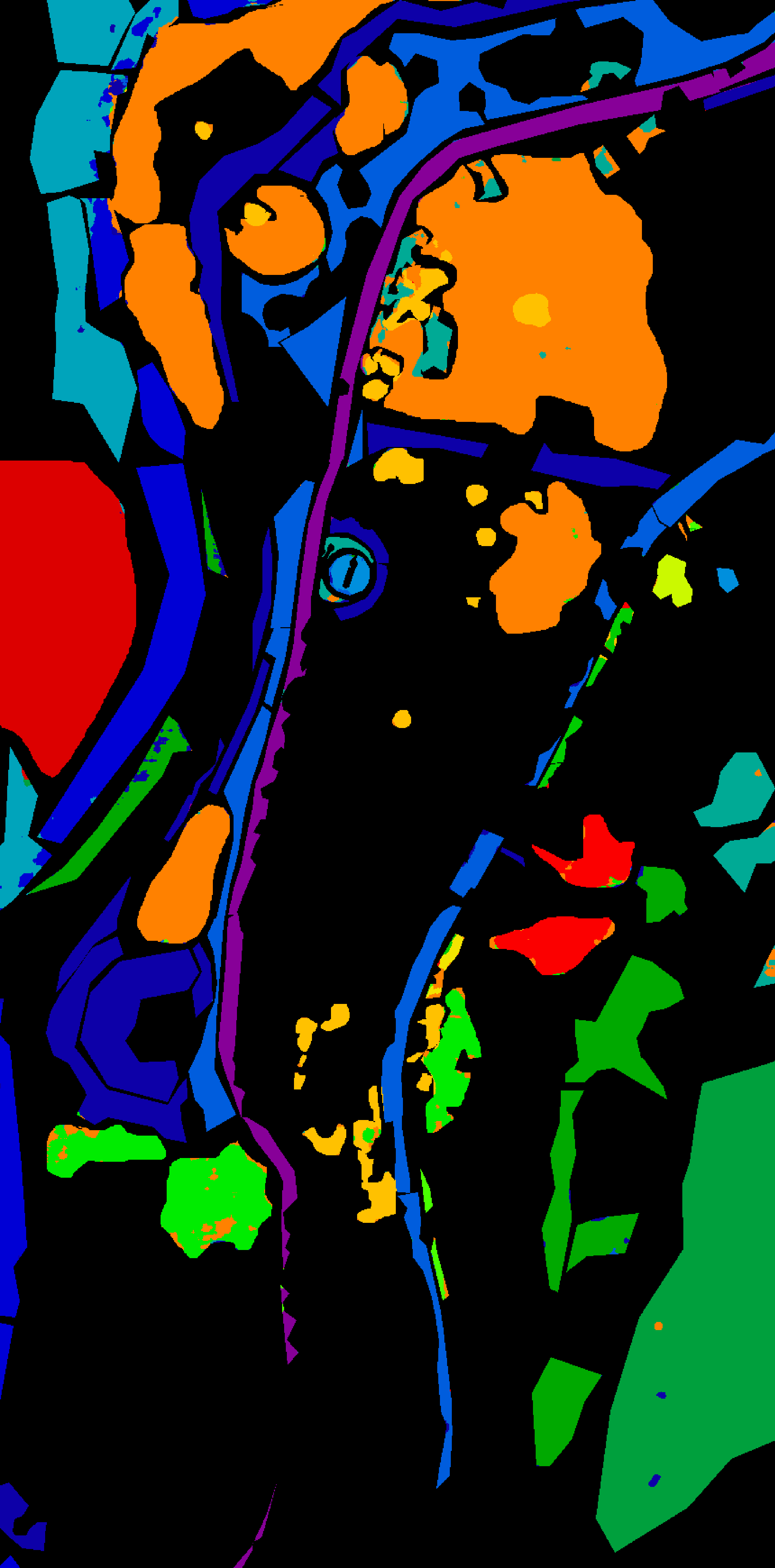}} \hspace{0.02cm}
    \subfloat[S2CIFT]{\includegraphics[width=0.13\columnwidth]{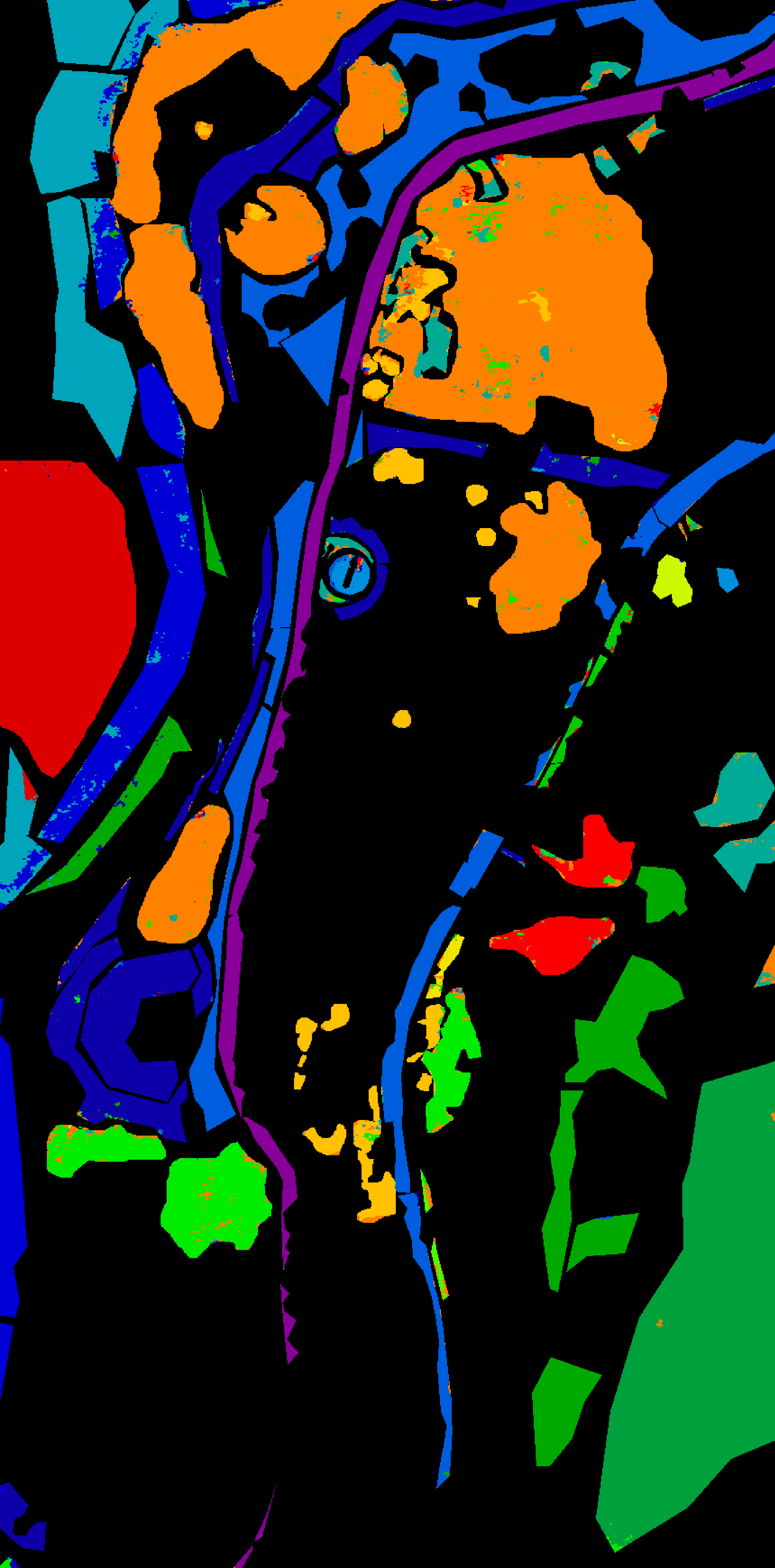}} \hspace{0.02cm}
    \subfloat[S2CAT]{\includegraphics[width=0.13\columnwidth]{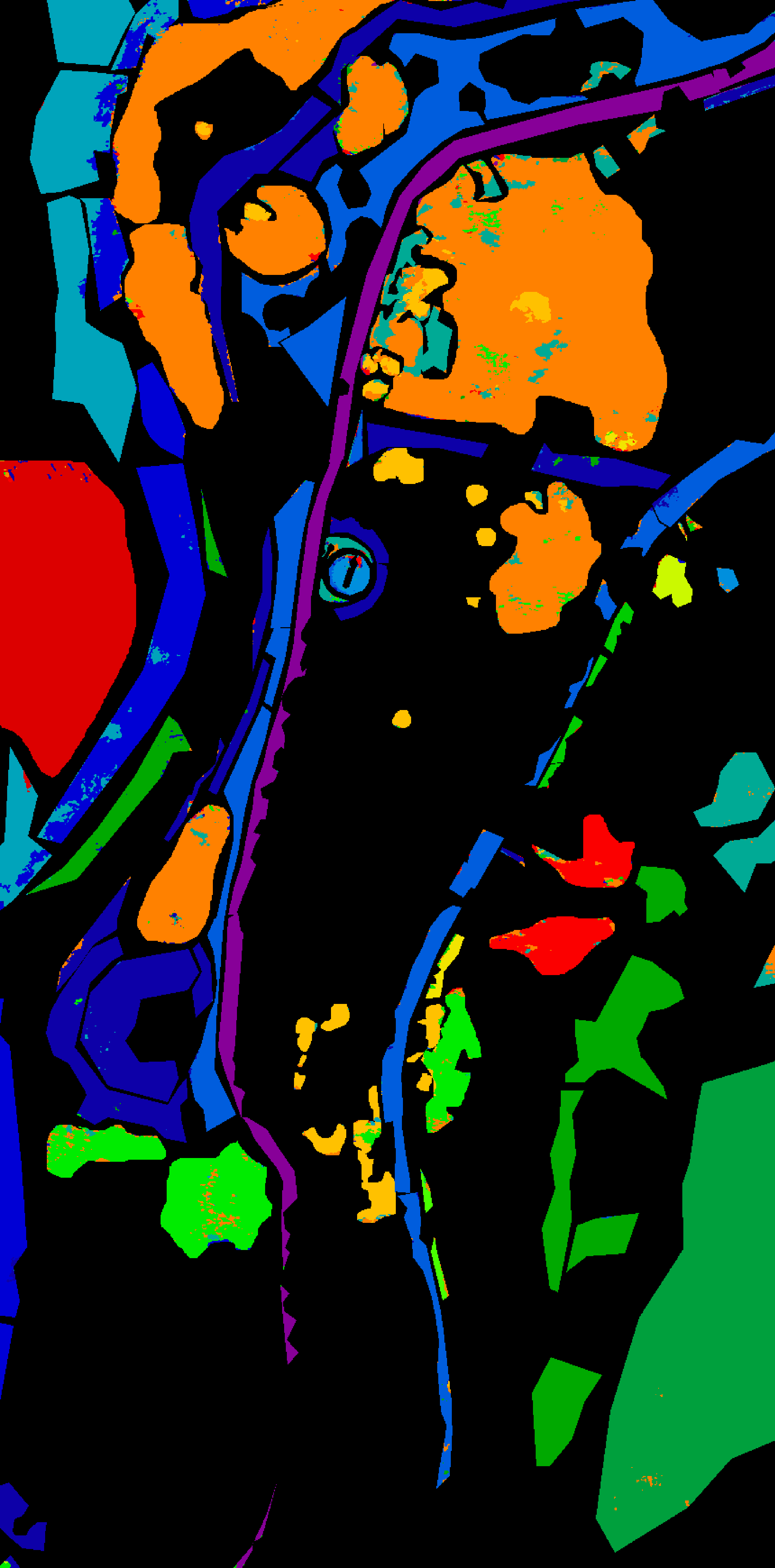}} \hspace{0.02cm}
    \subfloat[WF]{\includegraphics[width=0.13\columnwidth]{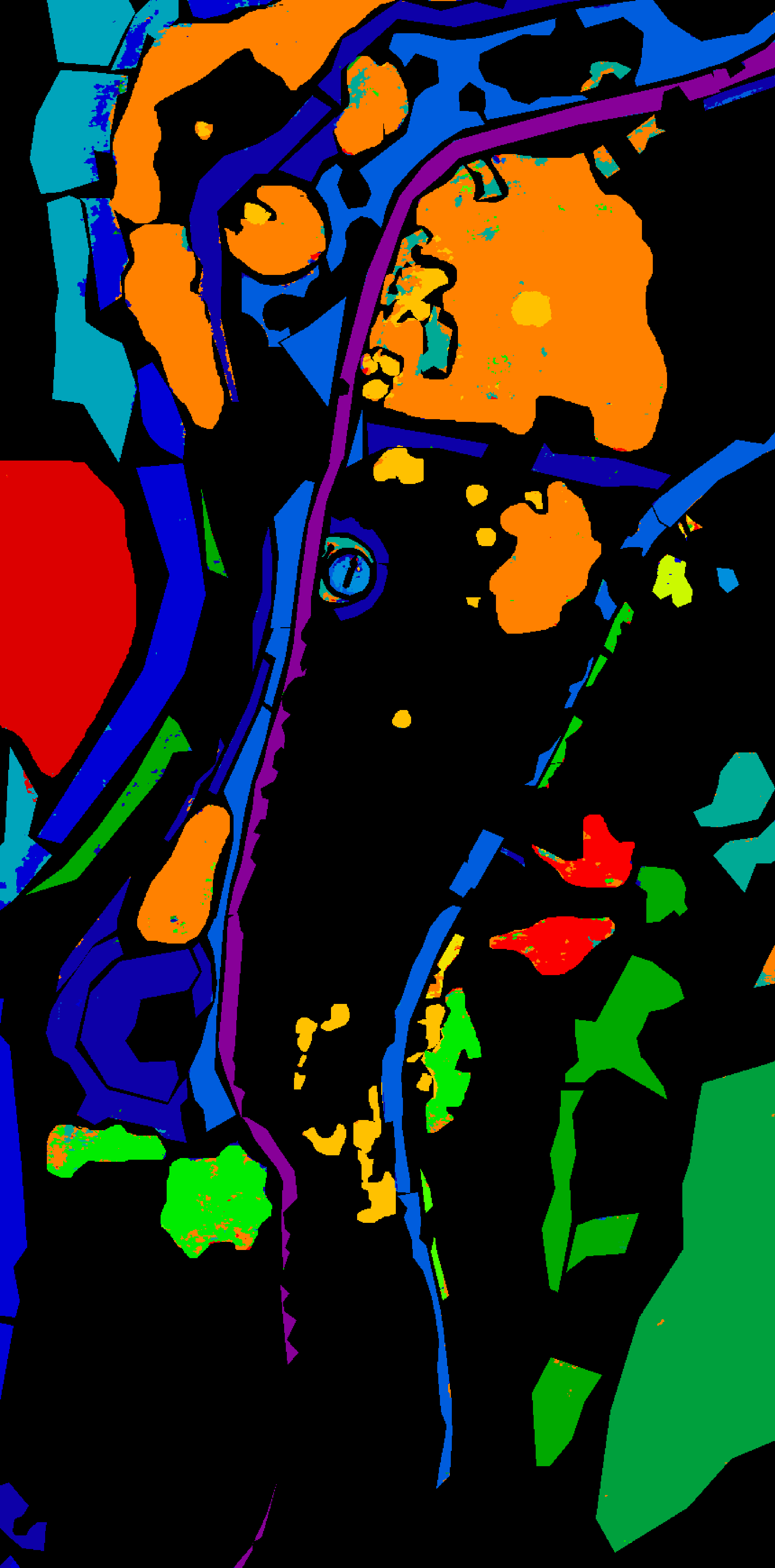}} \hspace{0.02cm}
    \subfloat[DF]{\includegraphics[width=0.13\columnwidth]{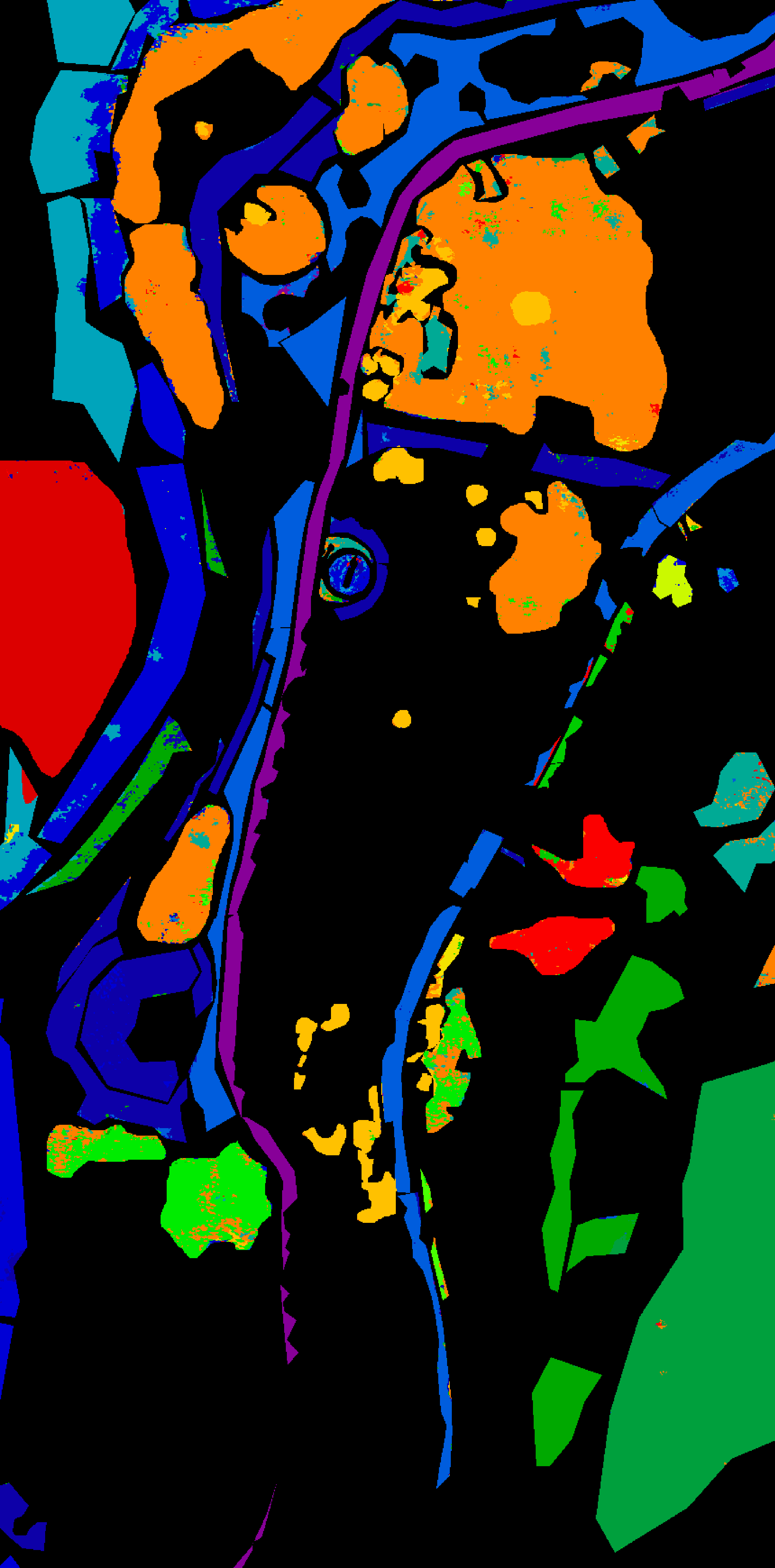}} \hspace{0.02cm}
    \subfloat[SSAM]{\includegraphics[width=0.13\columnwidth]{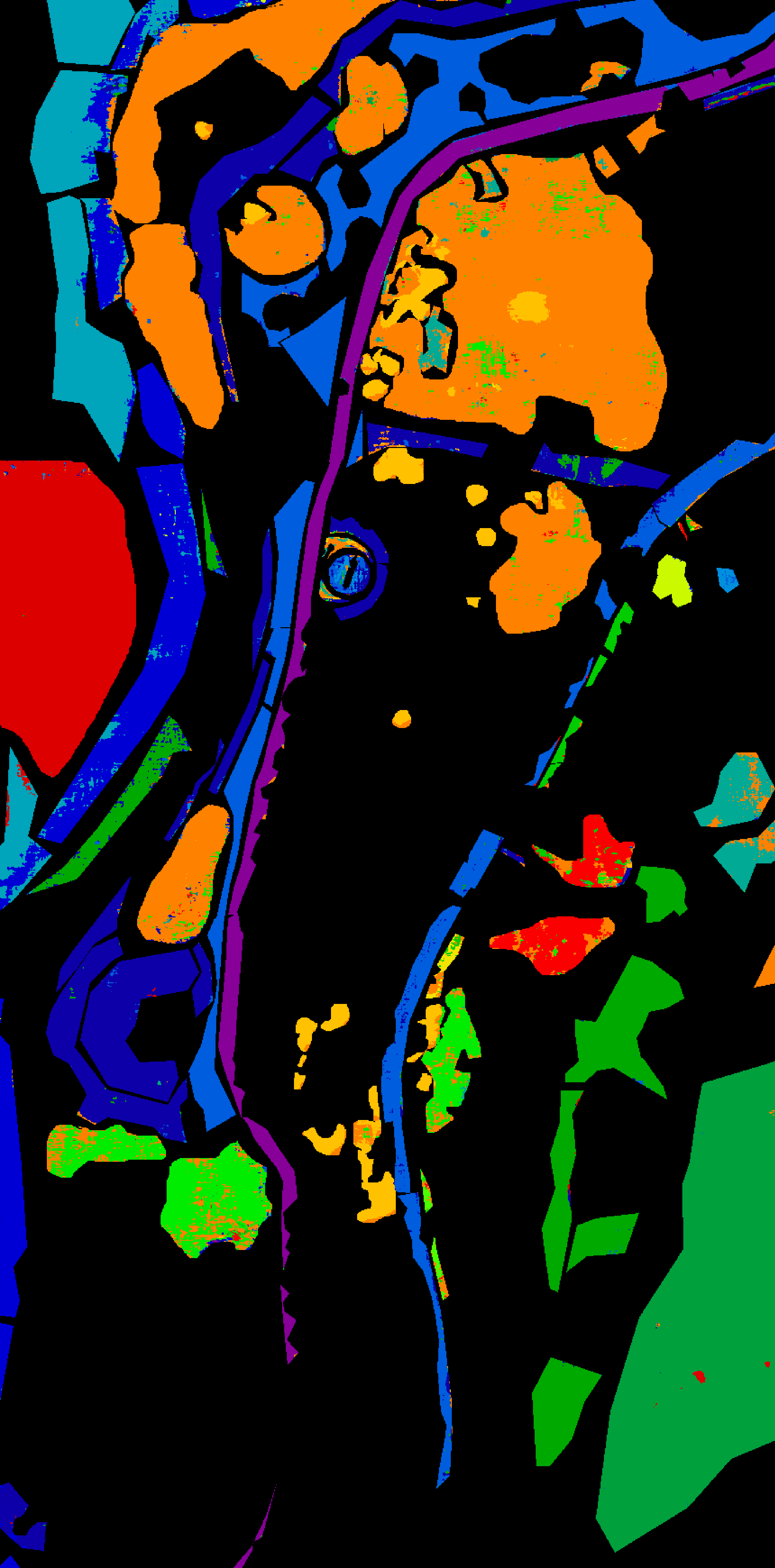}} \hspace{0.02cm}
    \subfloat[DBMLLA]{\includegraphics[width=0.13\columnwidth]{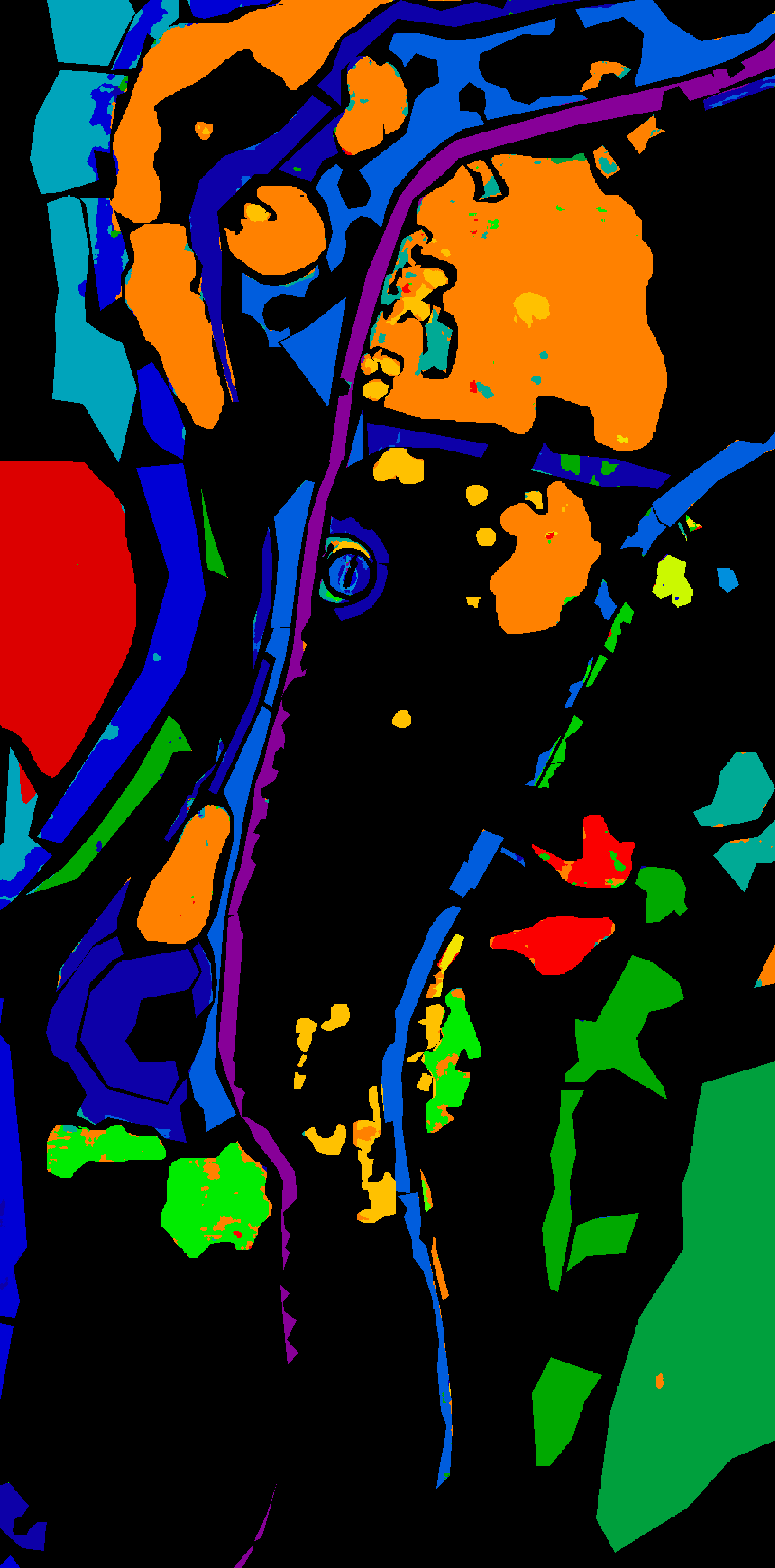}} \hspace{0.02cm}
    \subfloat[WDM]{\includegraphics[width=0.13\columnwidth]{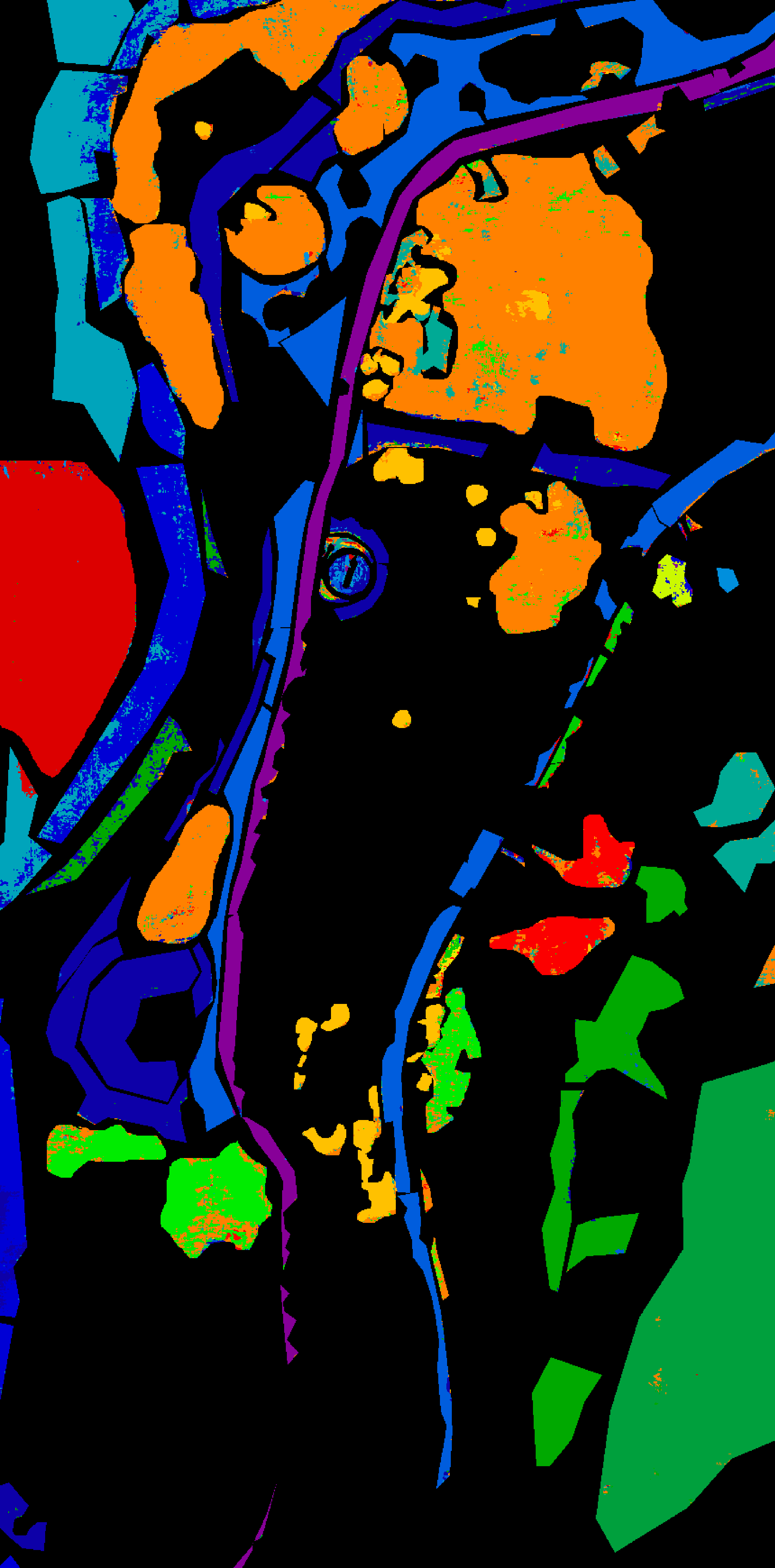}} \hspace{0.02cm}
    \subfloat[MM]{\includegraphics[width=0.13\columnwidth]{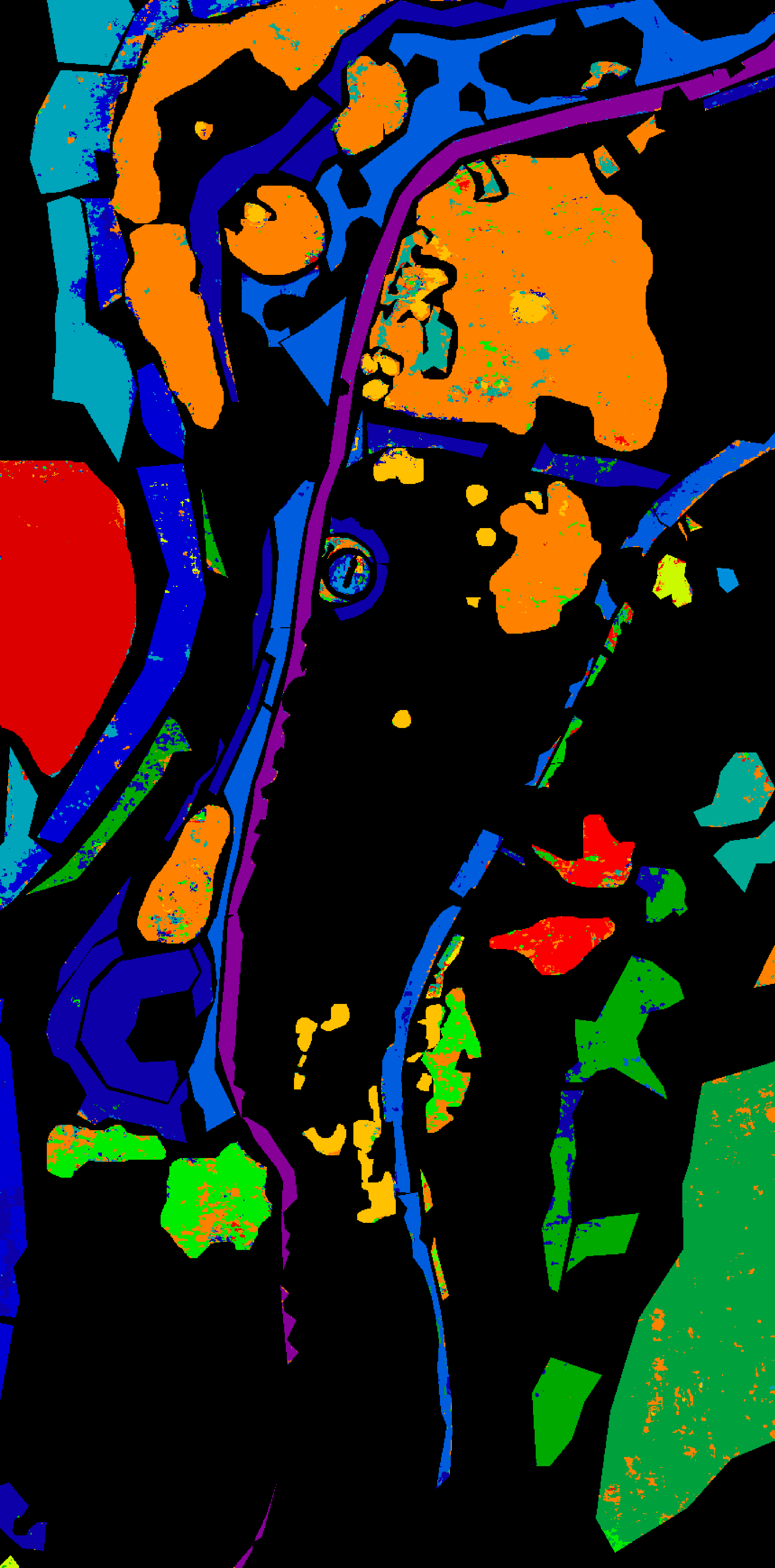}} \hspace{0.02cm}
    \subfloat[GM]{\includegraphics[width=0.13\columnwidth]{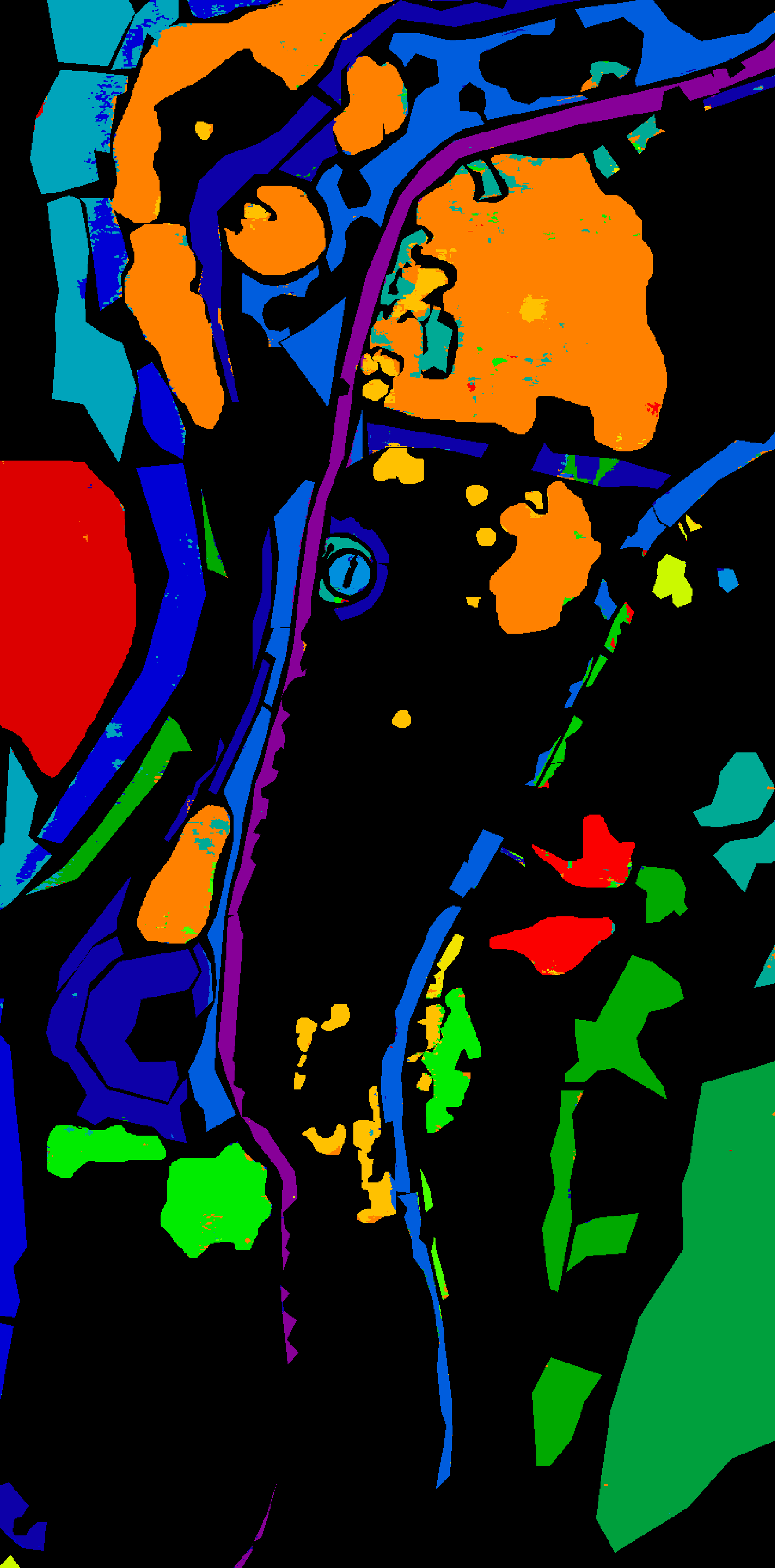}} \hspace{0.02cm}
    \subfloat[PM]{\includegraphics[width=0.13\columnwidth]{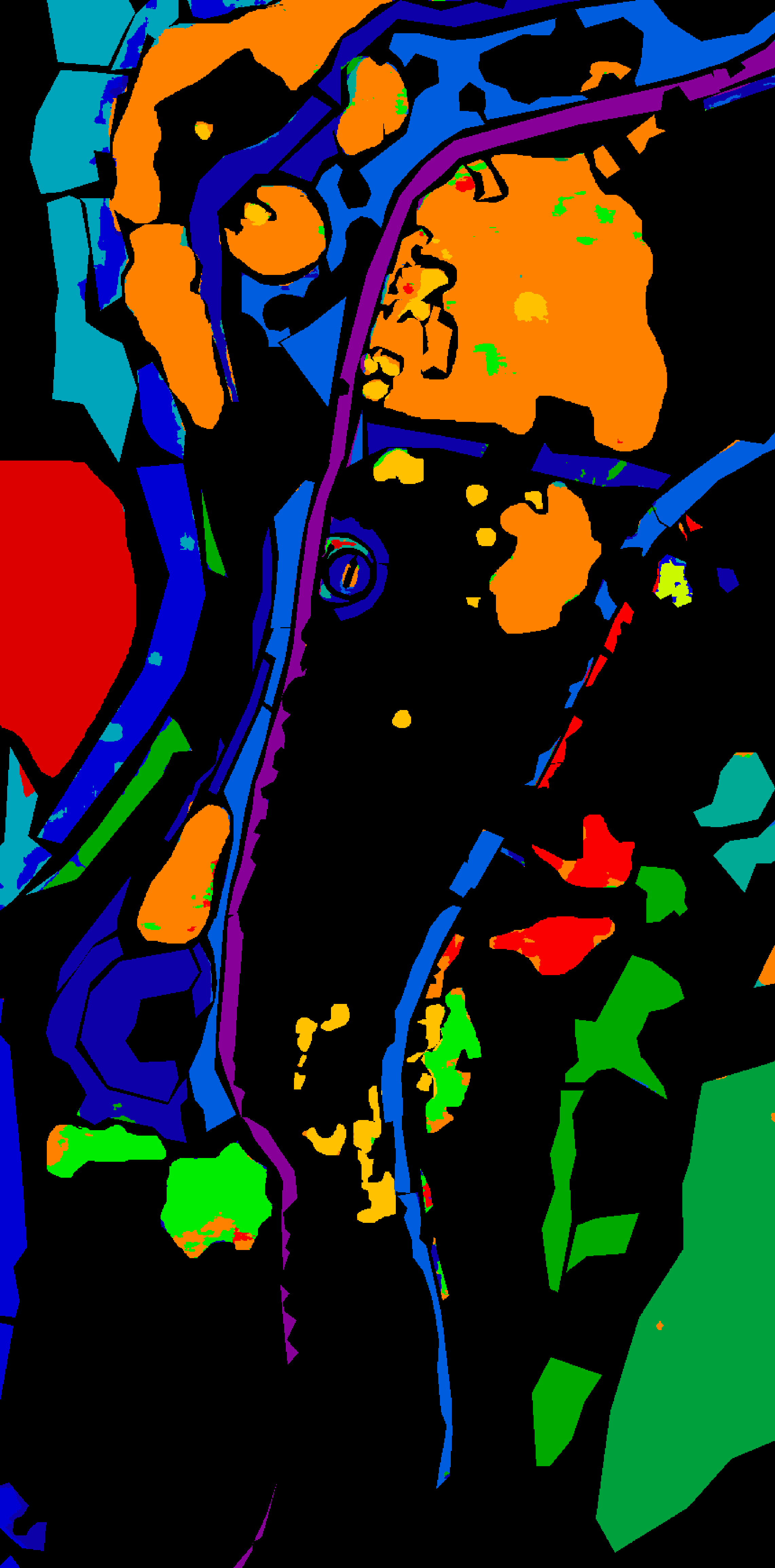}} \hspace{0.02cm}
    \subfloat[KDM]{\includegraphics[width=0.13\columnwidth]{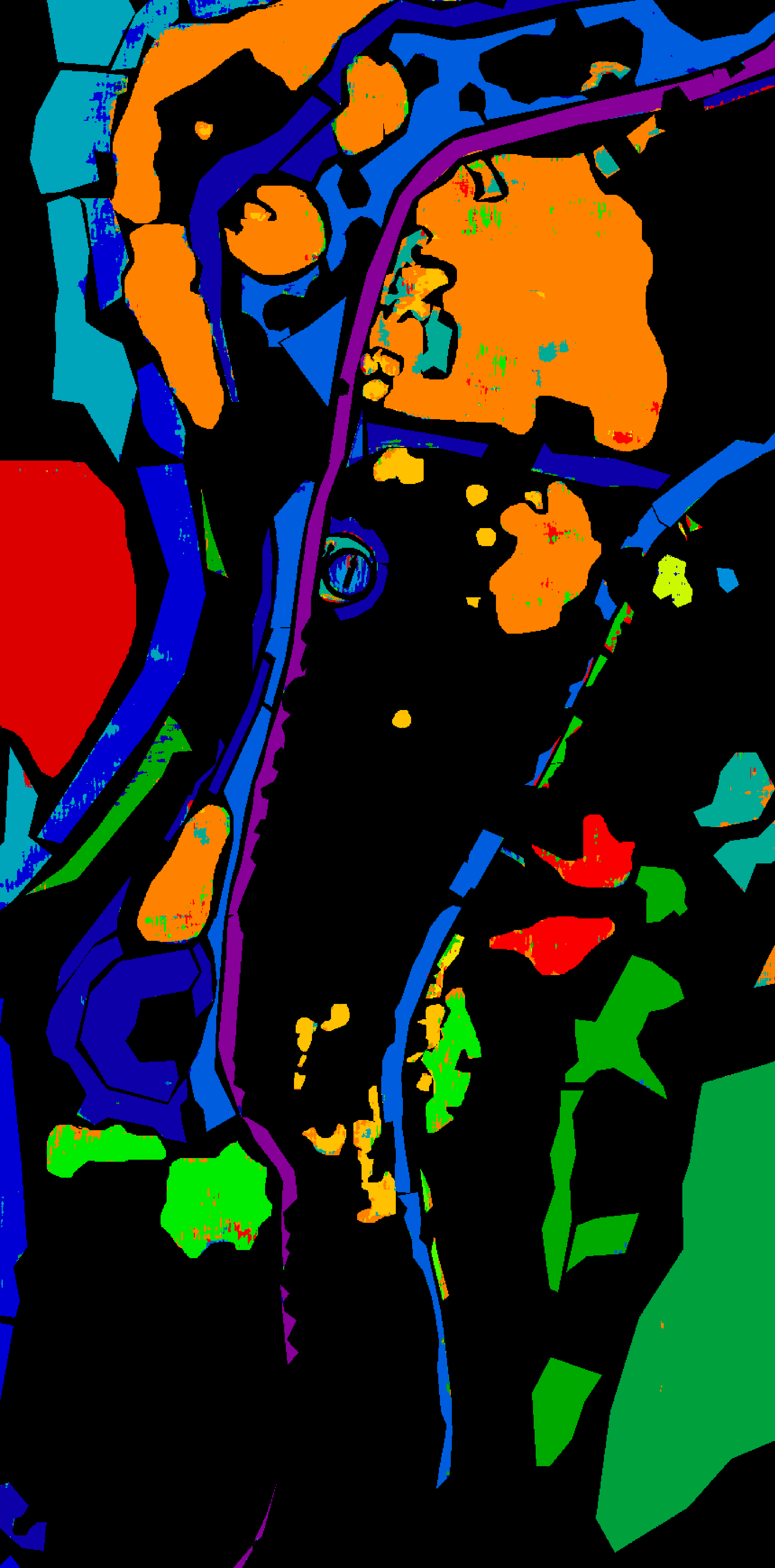}} \hspace{0.02cm}
    \subfloat[Pro.]{\includegraphics[width=0.13\columnwidth]{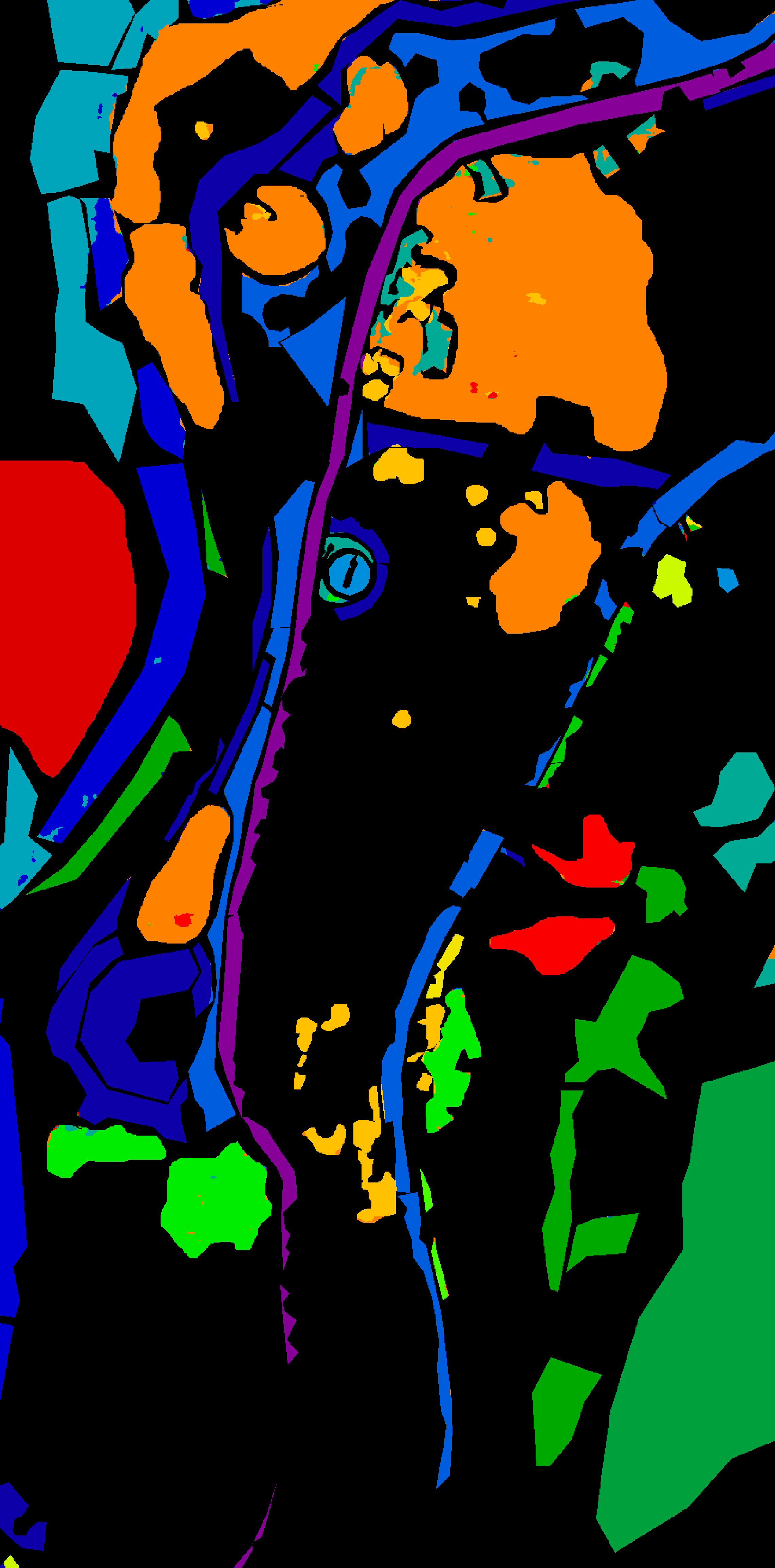}}
    \caption{Visual comparison of classification maps on TD.}
    \label{TDGT}
\end{figure}

\begin{figure}[!hbt]
    \centering
    \subfloat[GT]{\includegraphics[width=0.13\columnwidth]{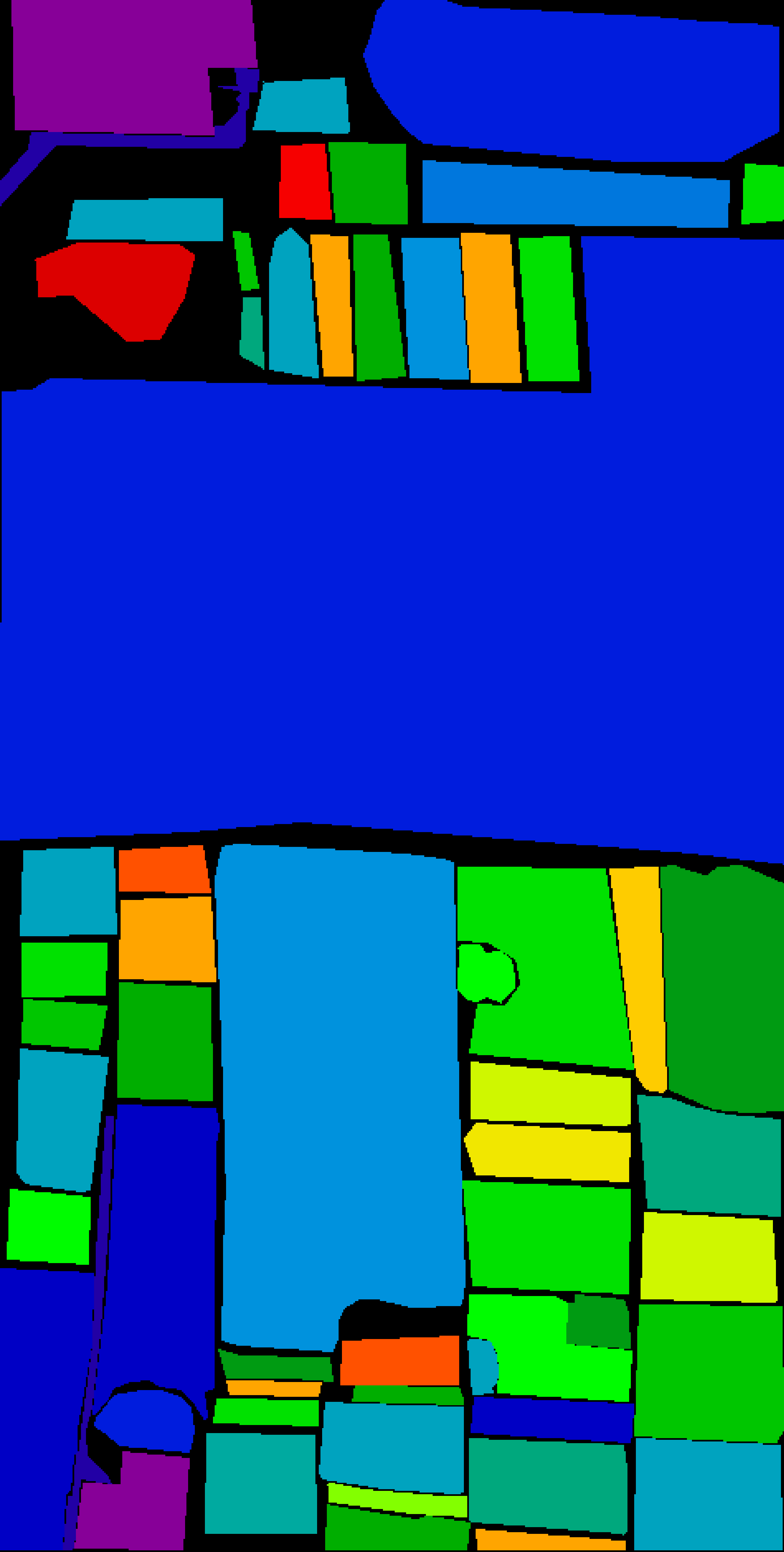}} \hspace{0.02cm}
    \subfloat[MSST]{\includegraphics[width=0.13\columnwidth]{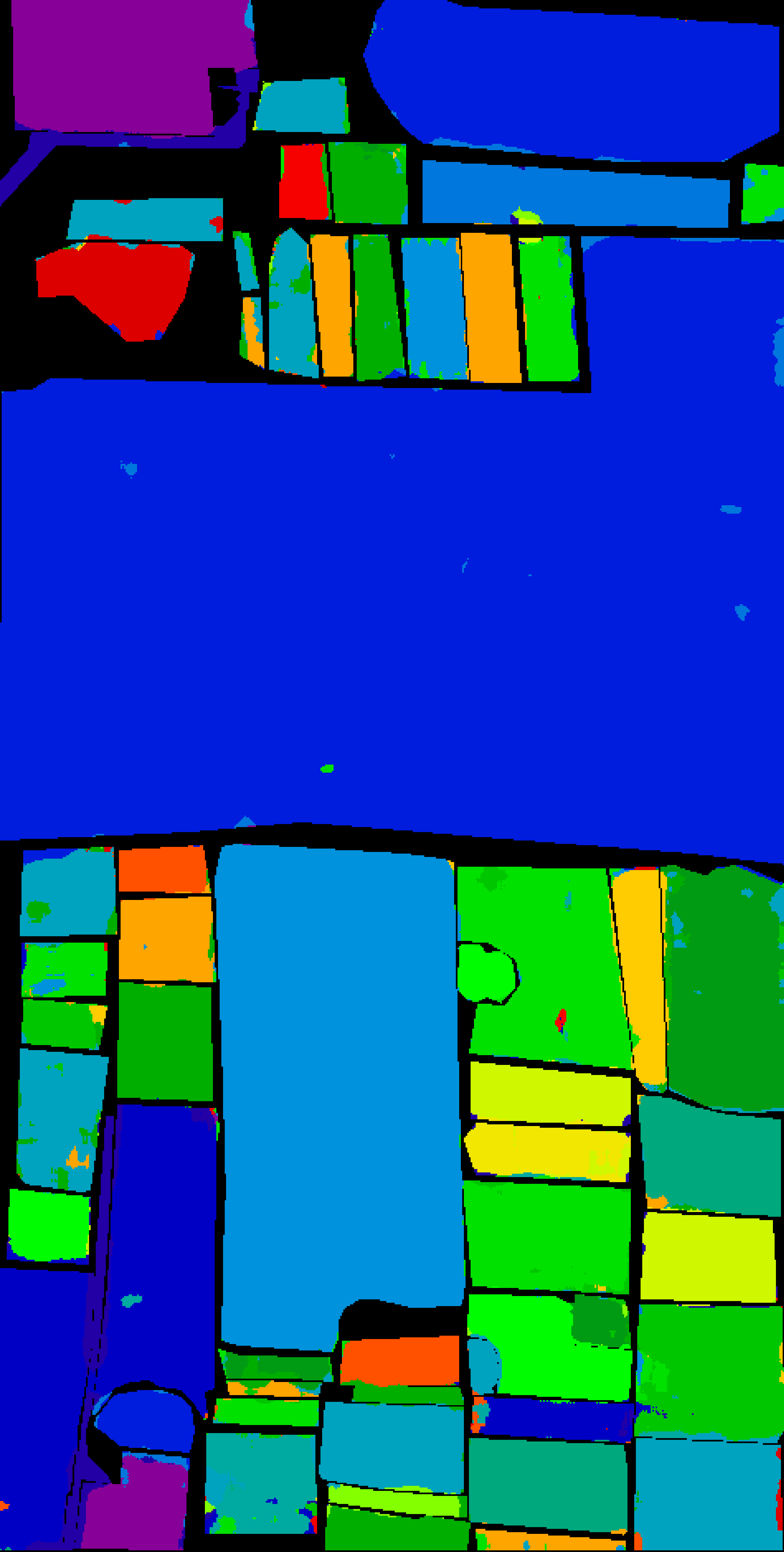}} \hspace{0.02cm}
    \subfloat[S2CIFT]{\includegraphics[width=0.13\columnwidth]{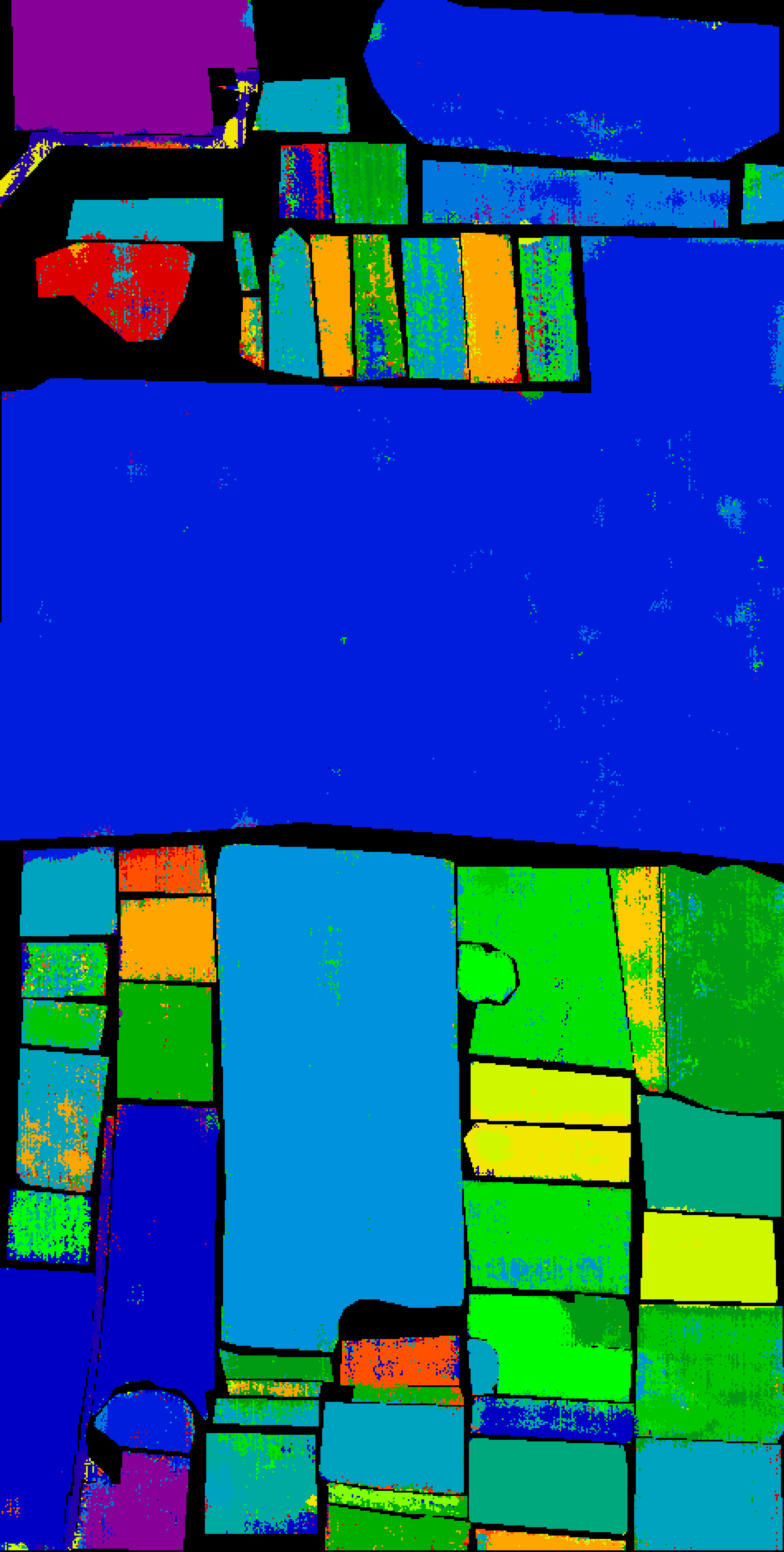}} \hspace{0.02cm}
    \subfloat[S2CAT]{\includegraphics[width=0.13\columnwidth]{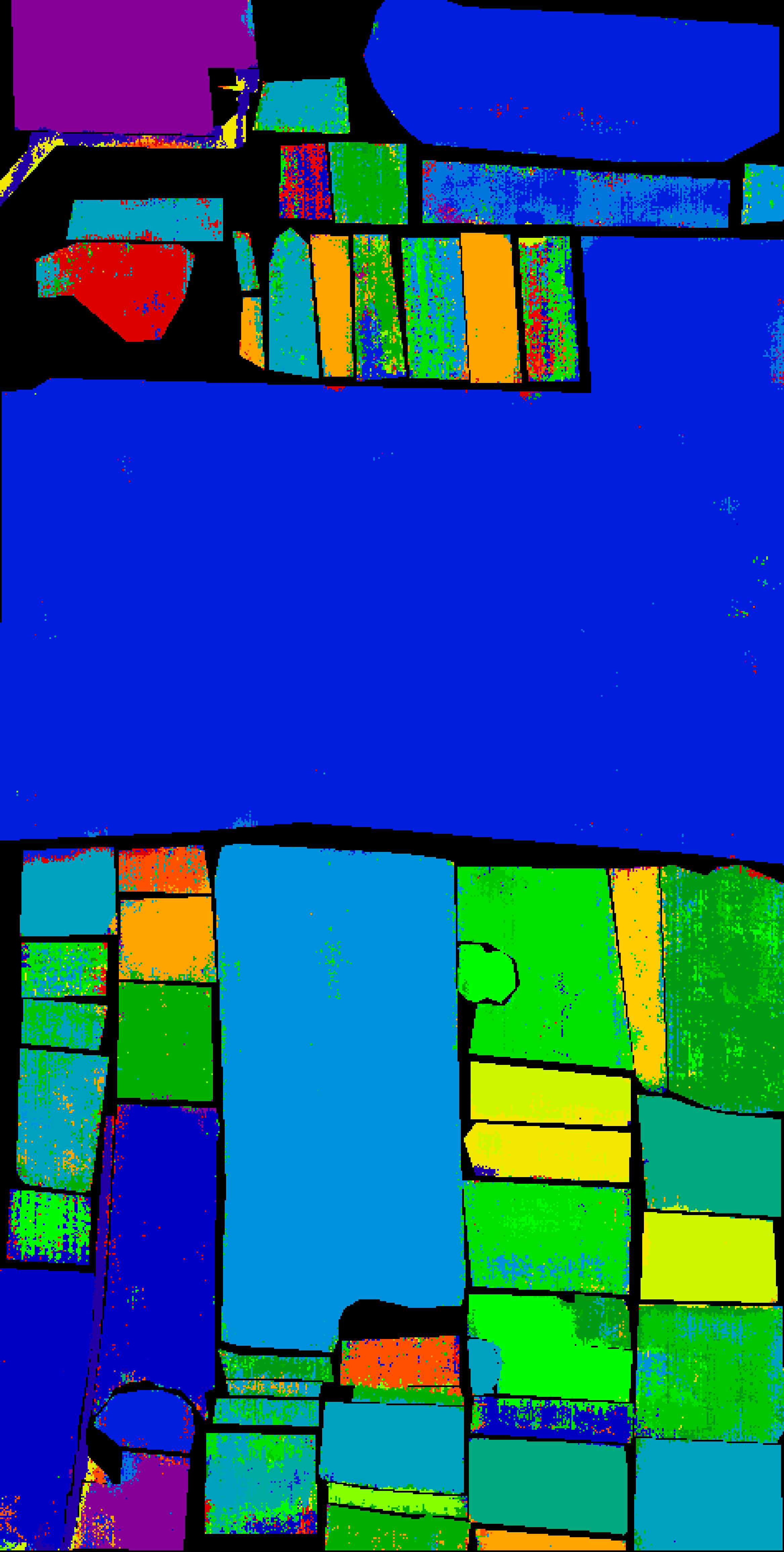}} \hspace{0.02cm}
    \subfloat[WF]{\includegraphics[width=0.13\columnwidth]{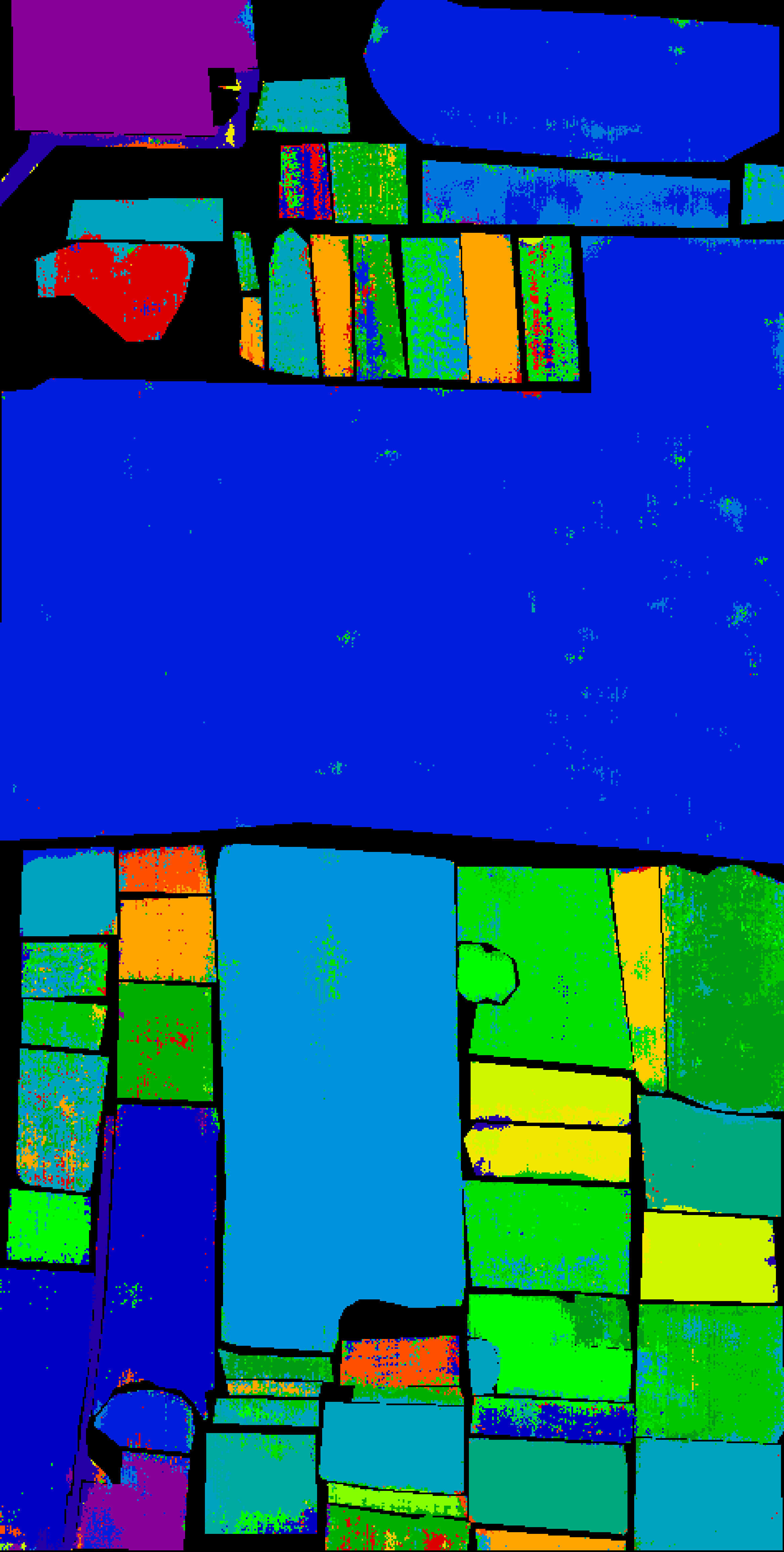}} \hspace{0.02cm}
    \subfloat[DF]{\includegraphics[width=0.13\columnwidth]{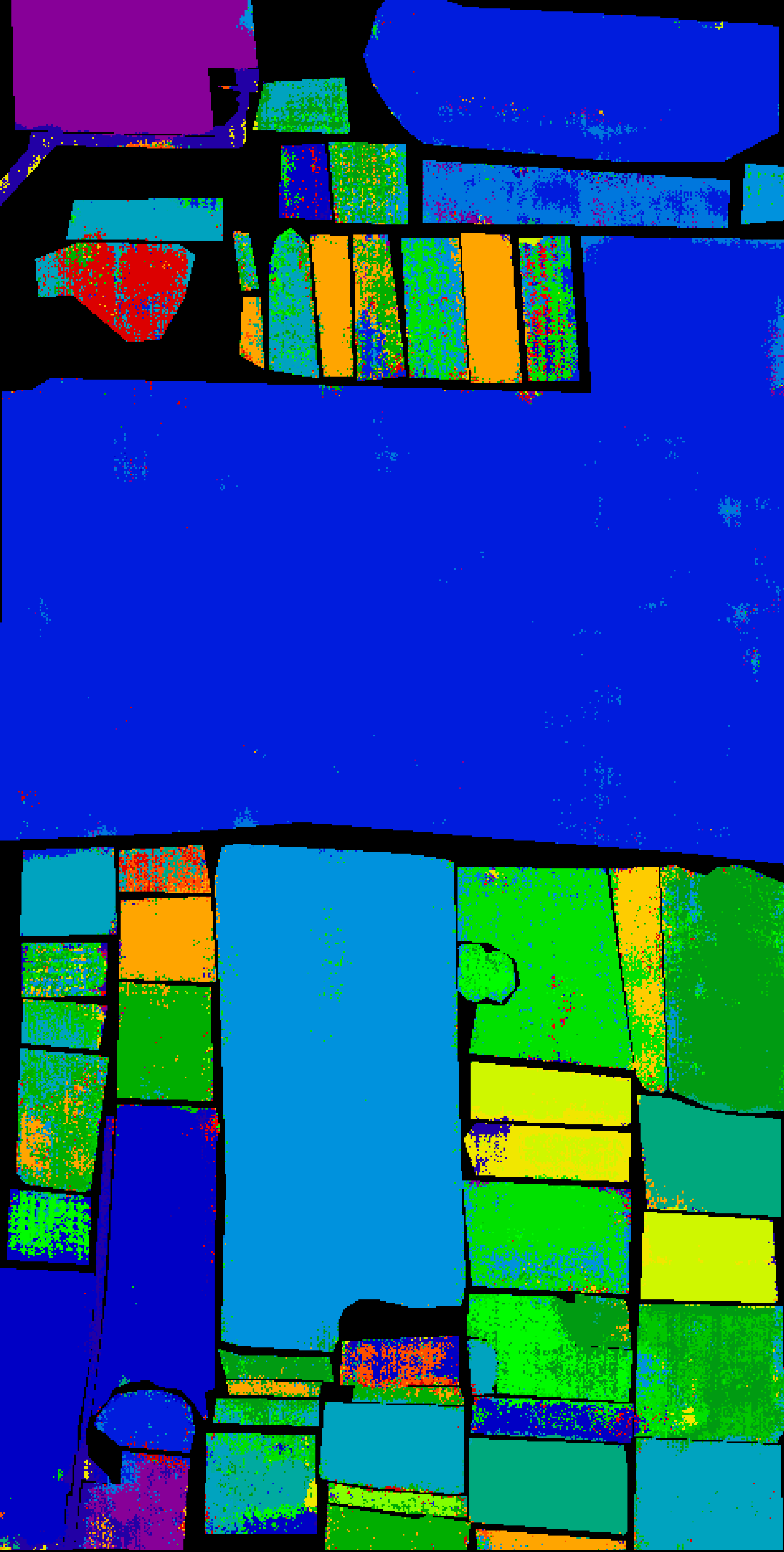}} \hspace{0.02cm}
    \subfloat[SSAM]{\includegraphics[width=0.13\columnwidth]{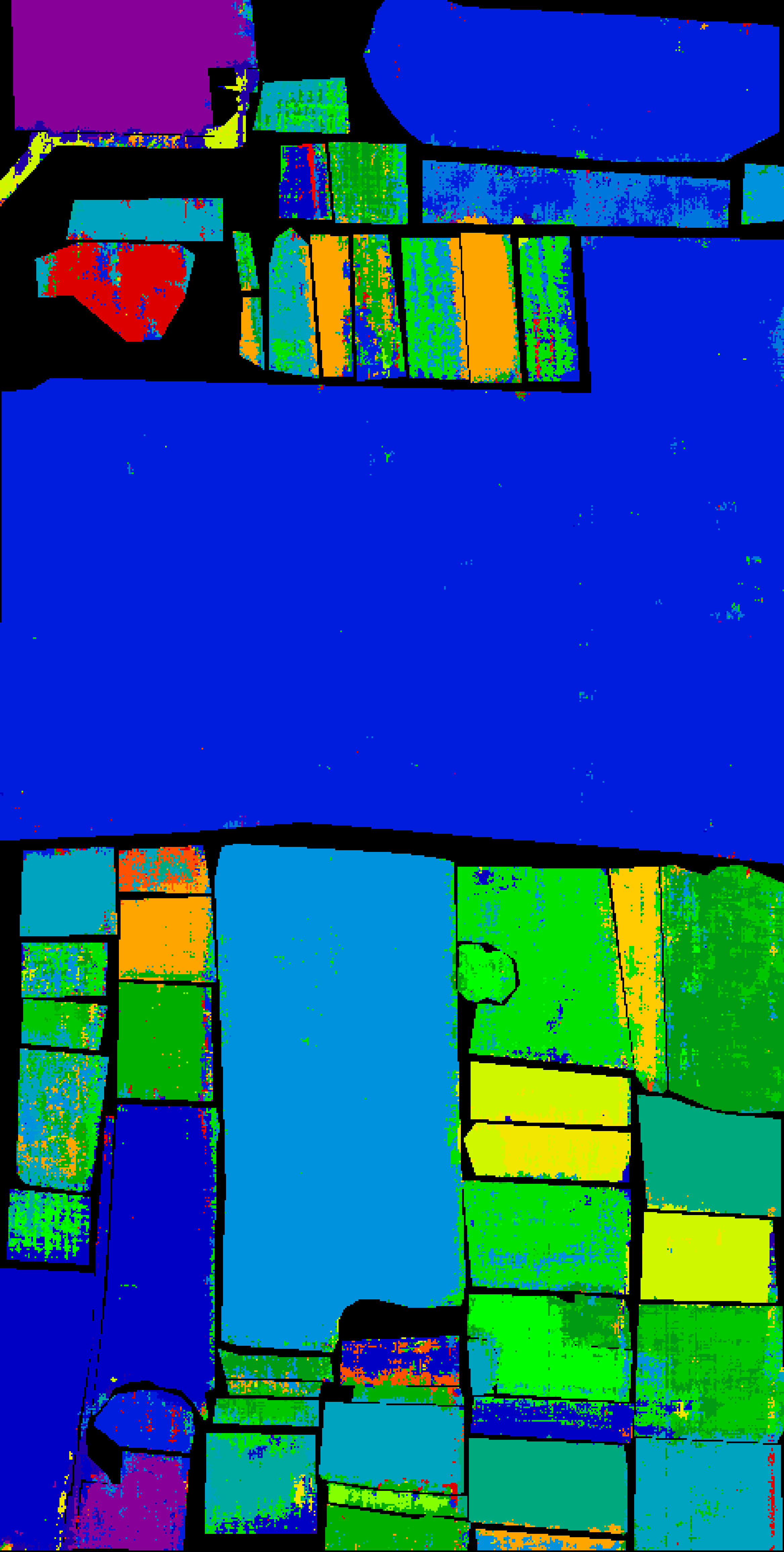}} \hspace{0.02cm}
    \subfloat[DBMLLA]{\includegraphics[width=0.13\columnwidth]{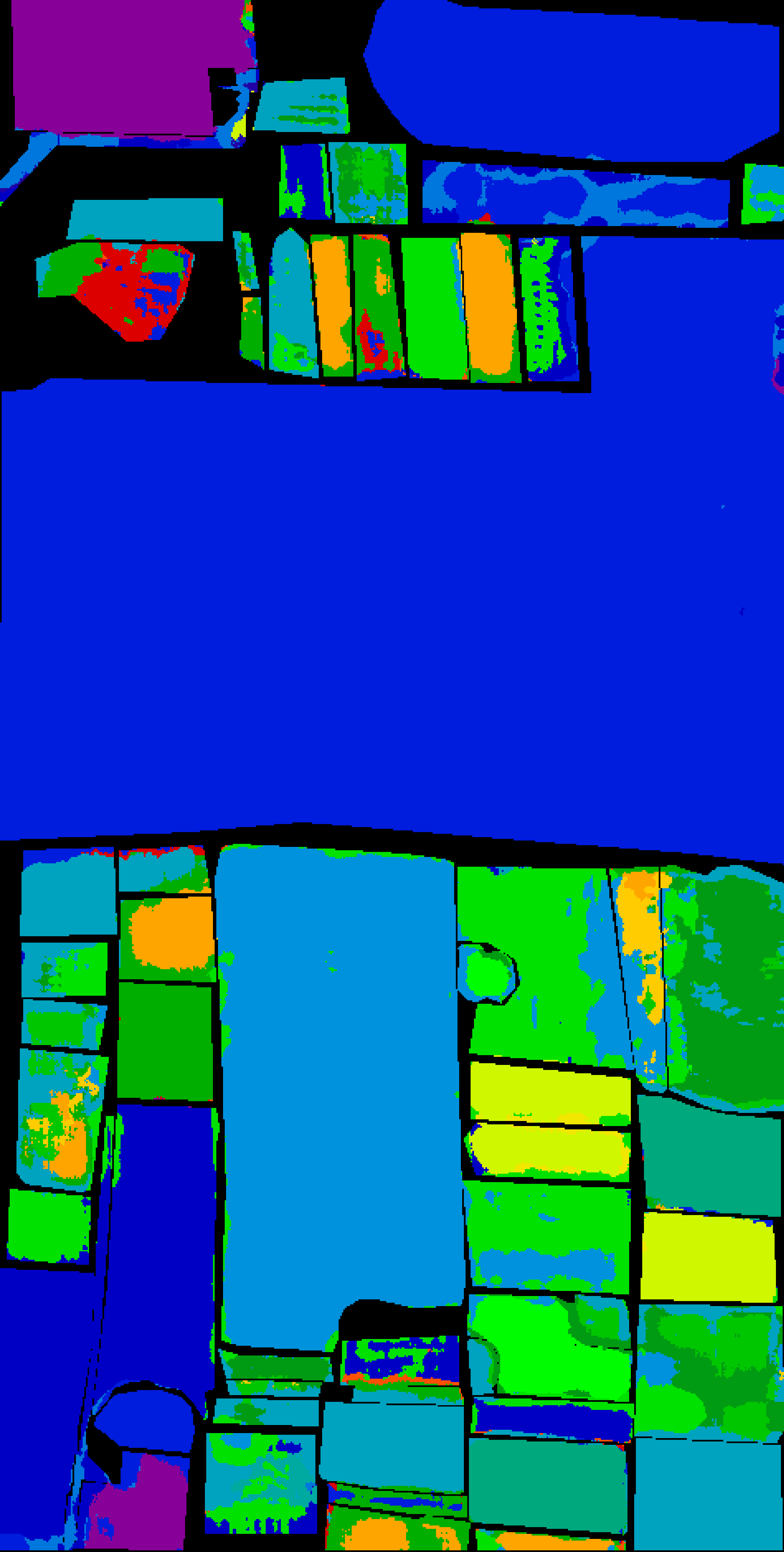}} \hspace{0.02cm}
    \subfloat[WDM]{\includegraphics[width=0.13\columnwidth]{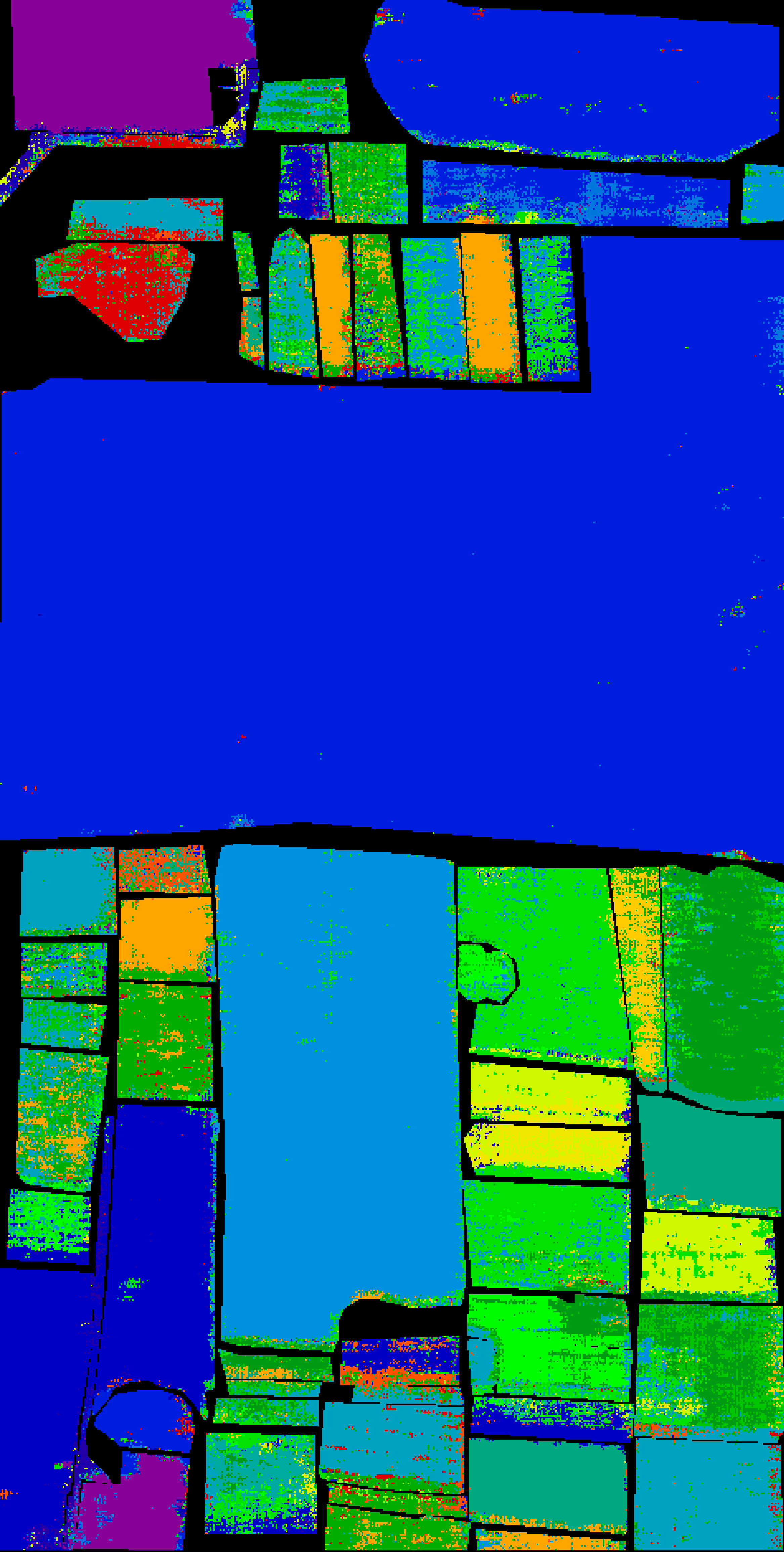}} \hspace{0.02cm}
    \subfloat[MM]{\includegraphics[width=0.13\columnwidth]{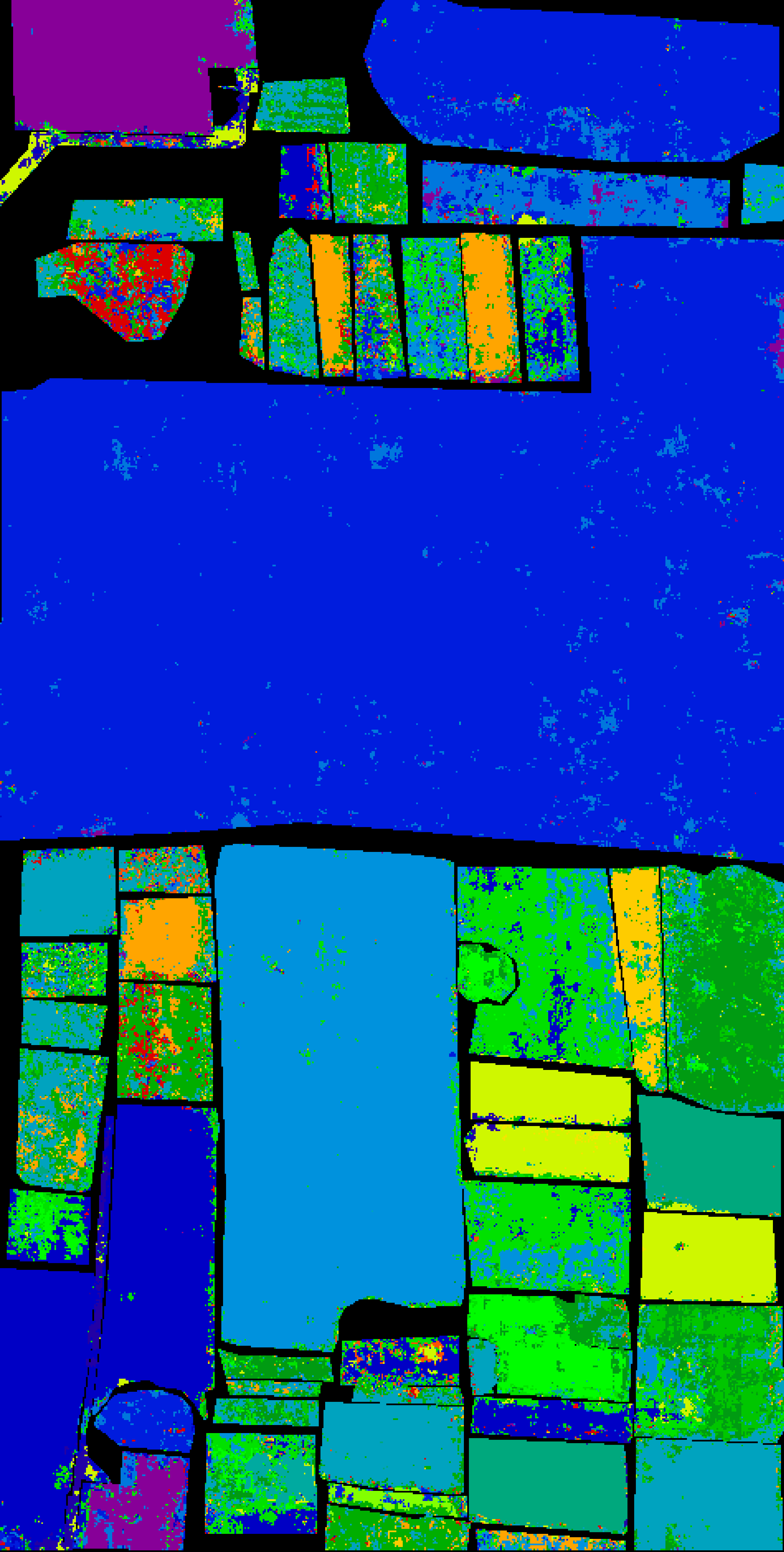}} \hspace{0.02cm}
    \subfloat[GM]{\includegraphics[width=0.13\columnwidth]{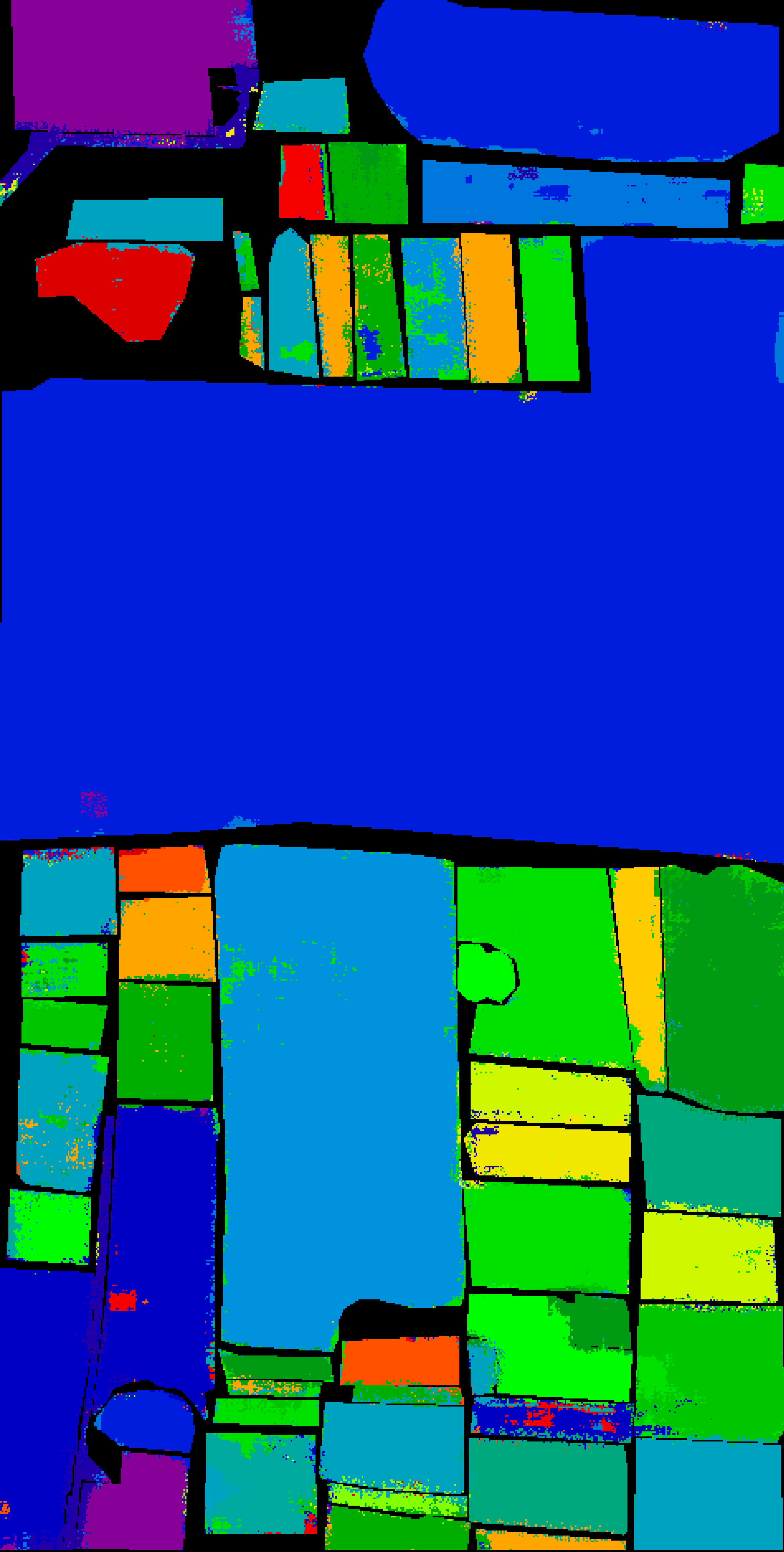}} \hspace{0.02cm}
    \subfloat[PM]{\includegraphics[width=0.13\columnwidth]{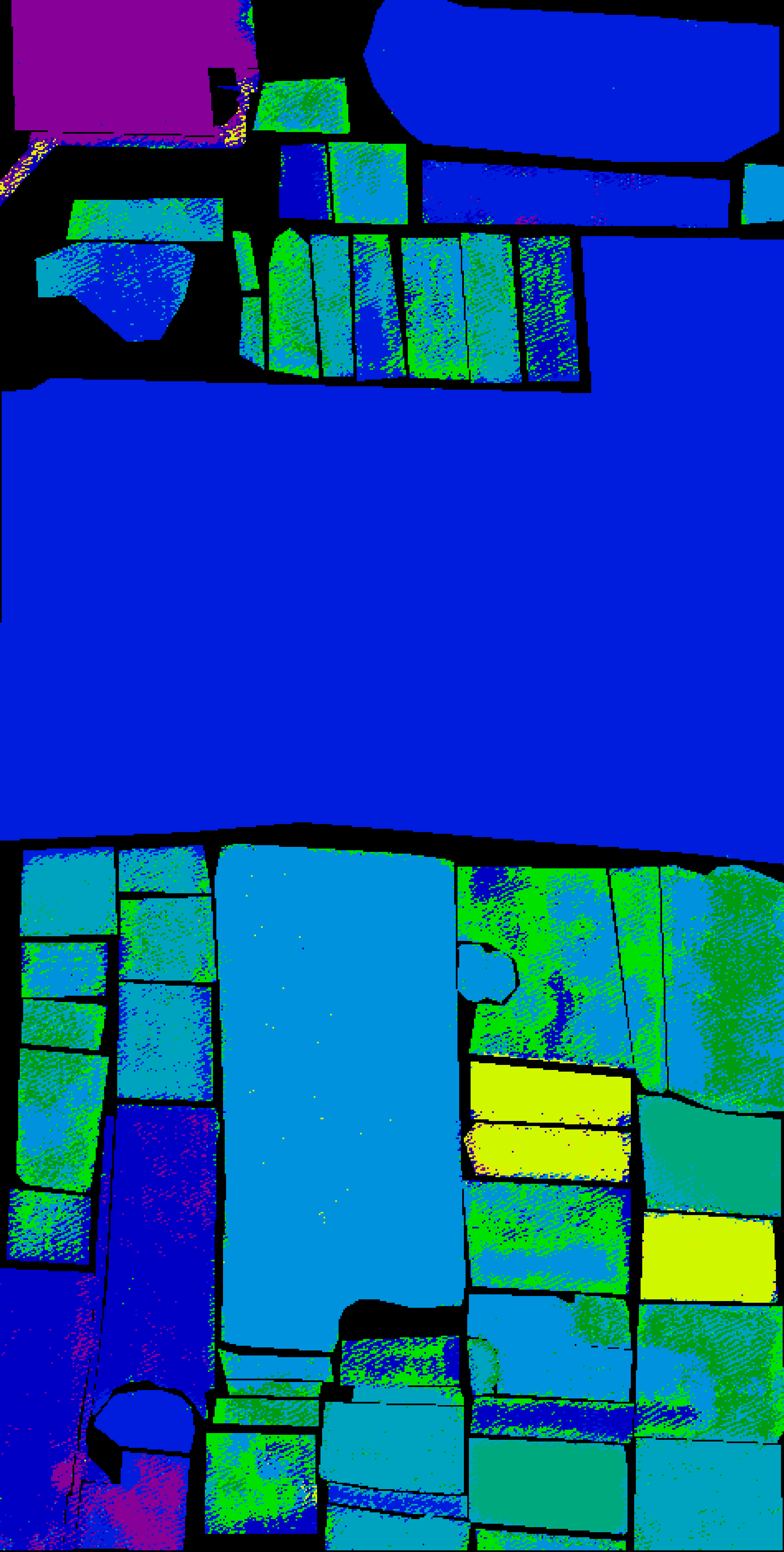}} \hspace{0.02cm}
    \subfloat[KDM]{\includegraphics[width=0.13\columnwidth]{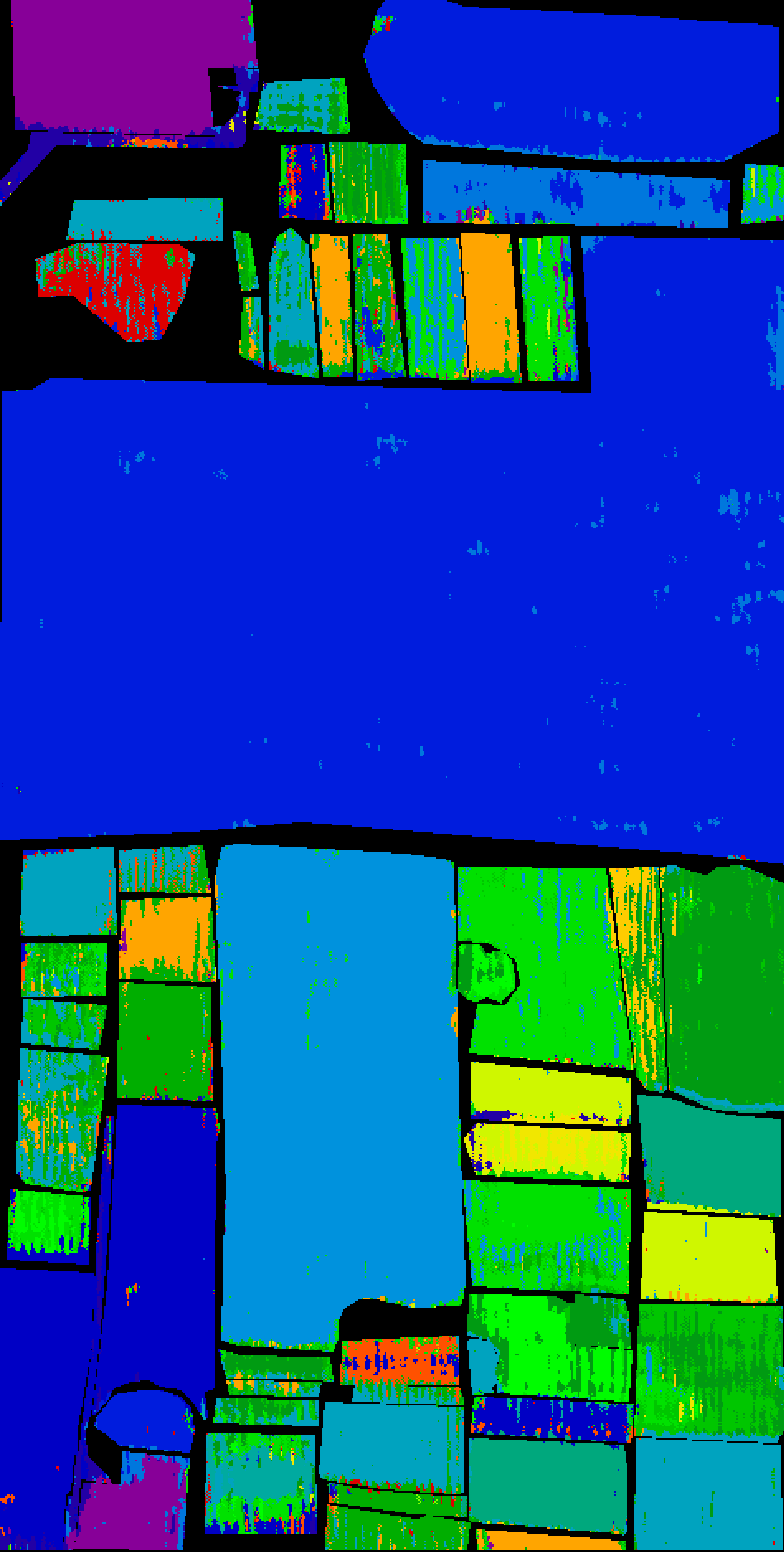}} \hspace{0.02cm}
    \subfloat[Pro.]{\includegraphics[width=0.13\columnwidth]{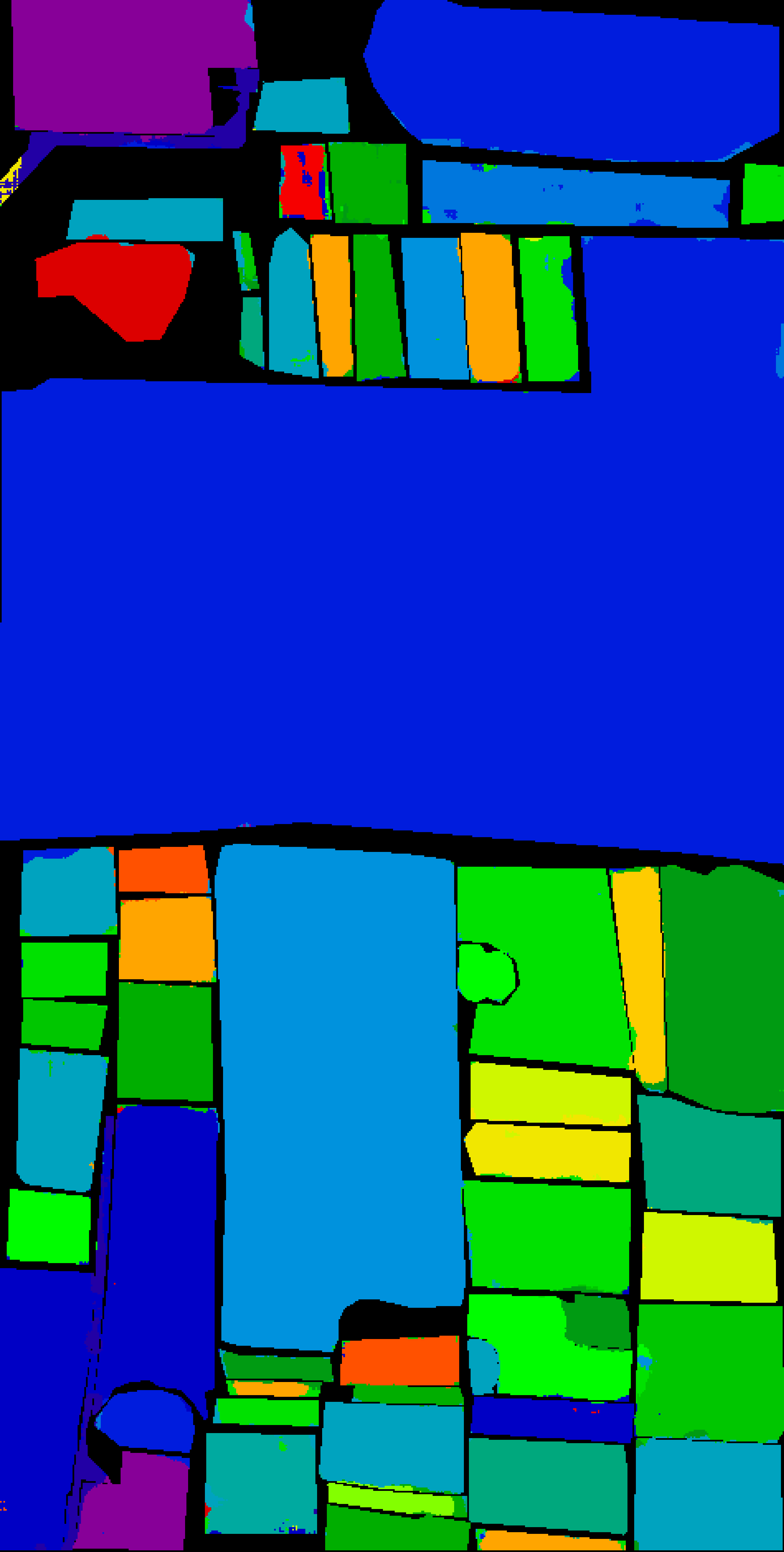}}
    \caption{Visual comparison of classification maps on HH.}
    \label{HHGT}
\end{figure}

Table~\ref{tab:placeholder} reports Kappa ($\kappa$), overall (OA), and average accuracy (AA) for the compared methods. Overall, the proposed cosine-normalized attention variants remain competitive with, and often outperform, recent CNN-, Transformer-, and Mamba-based baselines under the 1\% training protocol. This is notable because the experiments are conducted in a highly label-scarce regime where overfitting is a practical concern. Among the evaluated score formulations, normalized cosine-based variants perform consistently well. On TD, CS$^2$ achieves the best $\kappa$ and OA. On SA, C-SDP yields the best OA and $\kappa$, while MSA-CS$^2$ gives the best AA. On HH, CS attains the best $\kappa$ and OA, whereas Add gives the best AA. These results suggest that cosine-based scoring provides a reliable inductive bias for HSIC across different scenes.

A key observation is that methods relying on cosine-based similarity consistently rank among the top-performing variants across all datasets, despite sharing the same backbone with other attention formulations. This indicates that performance gains are primarily driven by the choice of similarity function rather than increased architectural complexity. In contrast, conventional dot-product and scaled dot-product attention exhibit slightly lower or less consistent performance, likely due to their sensitivity to feature magnitude variations. This effect is particularly relevant in hyperspectral data, where spectral magnitude can vary independently of material identity. By projecting features onto a unit hypersphere, cosine-normalized attention aligns the similarity computation with angular relationships, which are more discriminative for spectral signatures.

Furthermore, compared with recent Transformer- and Mamba-based models that introduce more complex architectural designs, the proposed approach achieves superior performance using a lightweight backbone. This suggests that improving the geometry of attention scoring can be as effective as increasing model complexity. Figs.~\ref{SAGT}-\ref{HHGT} show qualitative classification maps for representative methods. Compared with the baselines, the proposed model generally produces more homogeneous regions, fewer isolated errors, and cleaner class boundaries, especially in spectrally confusing areas. This behavior is consistent across datasets with different characteristics, indicating that the benefit of cosine-based scoring is not dataset-specific but arises from a more appropriate alignment between attention geometry and HSI properties.

\begin{figure}[!hbt]
    \centering
    \subfloat[SA]{\includegraphics[width=0.15\textwidth]{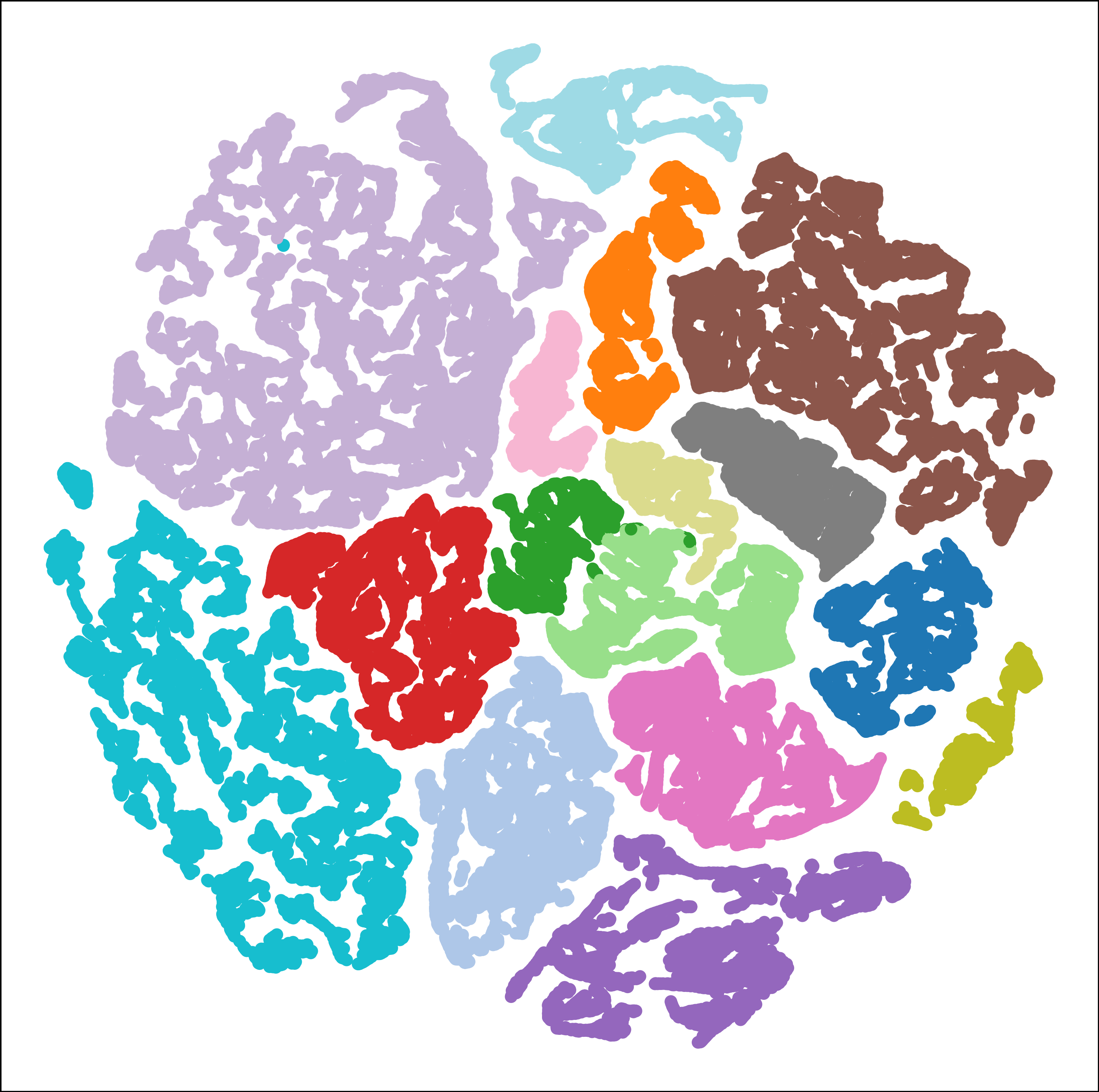}} \hspace{0.1cm}
    \subfloat[HH]{\includegraphics[width=0.15\textwidth]{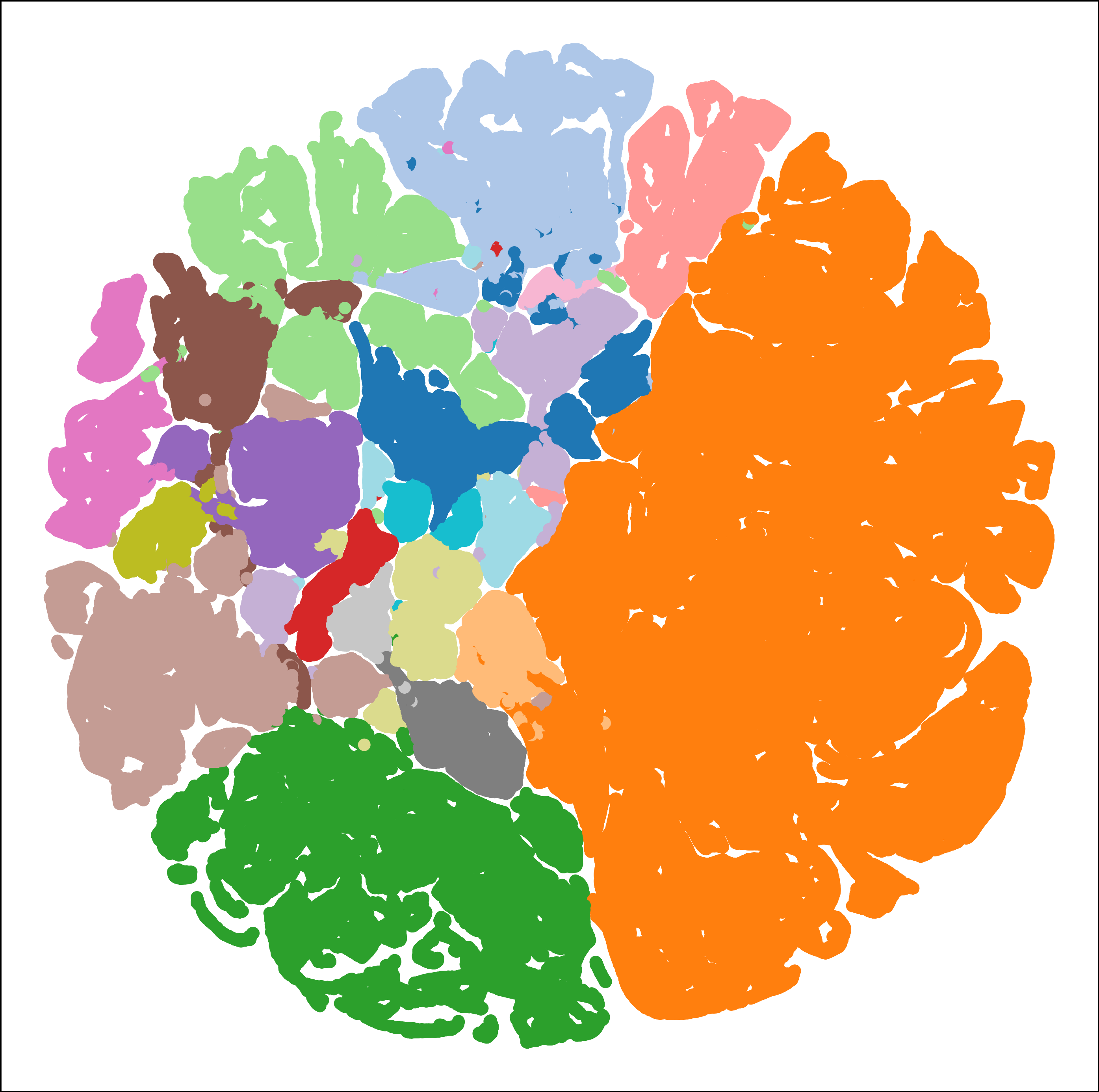}} \hspace{0.1cm}
    \subfloat[TD]{\includegraphics[width=0.15\textwidth]{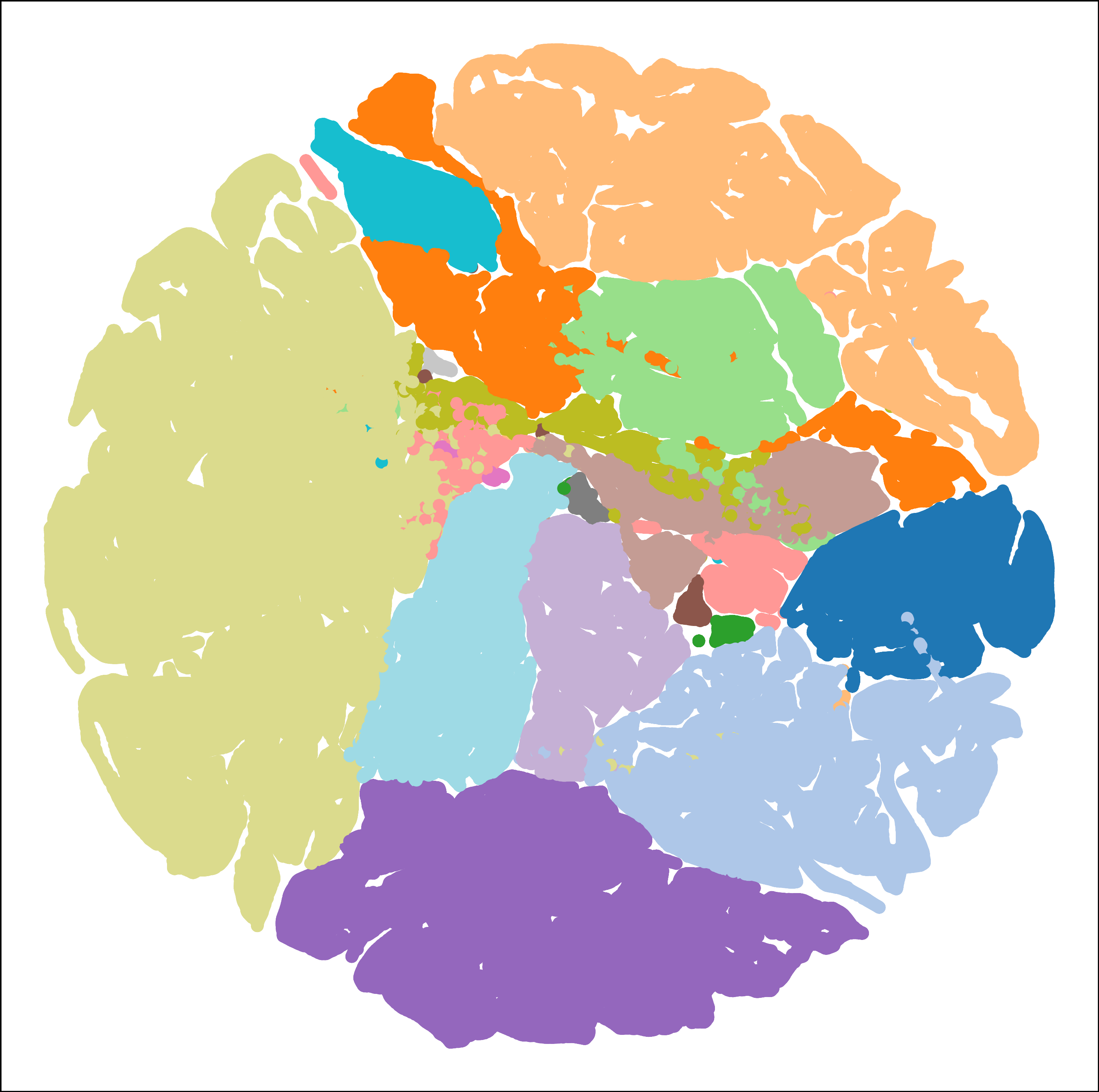}}
    \caption{t-SNE visualization of the learned feature spaces.}
    \label{fig:tSNE}
\end{figure}

\noindent\textbf{Feature-Space Visualization:} Fig.~\ref{fig:tSNE} visualizes the learned feature space using t-SNE on the final latent representations. Across all datasets, the embeddings form relatively compact and well-separated class clusters, indicating that the proposed attention learns discriminative spatial-spectral representations.

\noindent\textbf{Ablation and Robustness Analysis:} To provide a compact yet comprehensive analysis of the proposed framework, we report a unified study covering architectural choices, attention score design, patch-size sensitivity, and robustness to spectral noise. In each case, only the target factor is changed while the remaining settings are kept fixed. The results are reported in terms of OA (\%) on the SA, HH, and TD datasets.

\begin{table}[!hbt]
    \centering
    \caption{Ablation and robustness analysis.}
    \resizebox{\columnwidth}{!}{\begin{tabular}{l|l|ccc} \hline
        \textbf{Analysis Group} & \textbf{Configuration} & \textbf{SA} & \textbf{HH} & \textbf{TD} \\ \hline

        \multirow{4}{*}{Normalization}
        & None & 98.11 $\pm$ 0.12 & 97.64 $\pm$ 0.15 & 98.42 $\pm$ 0.11 \\
        & Query only & 98.74 $\pm$ 0.09 & 97.93 $\pm$ 0.12 & 98.58 $\pm$ 0.10 \\
        & Key only & 98.69 $\pm$ 0.10 & 97.89 $\pm$ 0.13 & 98.55 $\pm$ 0.09 \\
        & Query + Key & \textbf{99.24 $\pm$ 0.05} & \textbf{98.32 $\pm$ 0.07} & \textbf{98.84 $\pm$ 0.06} \\ \hline

        \multirow{4}{*}{Squaring strategy}
        & Cosine & 99.07 $\pm$ 0.07 & 98.21 $\pm$ 0.09 & 98.71 $\pm$ 0.08 \\
        & $|\mathrm{cosine}|$ & 99.11 $\pm$ 0.06 & 98.24 $\pm$ 0.08 & 98.74 $\pm$ 0.07 \\
        & Cosine$^{2}$ & \textbf{99.24 $\pm$ 0.05} & \textbf{98.32 $\pm$ 0.07} & \textbf{98.84 $\pm$ 0.06} \\
        & Temp.-scaled Cosine$^{2}$ & 99.19 $\pm$ 0.06 & 98.28 $\pm$ 0.08 & 98.79 $\pm$ 0.07 \\ \hline

        \multirow{3}{*}{Positional encoding}
        & None & 98.63 $\pm$ 0.11 & 97.71 $\pm$ 0.14 & 98.33 $\pm$ 0.12 \\
        & Sinusoidal & 99.12 $\pm$ 0.07 & 98.11 $\pm$ 0.09 & 98.68 $\pm$ 0.08 \\
        & Learnable & \textbf{99.24 $\pm$ 0.05} & \textbf{98.32 $\pm$ 0.07} & \textbf{98.84 $\pm$ 0.06} \\ \hline

        \multirow{3}{*}{Head number}
        & 2 heads & 99.08 $\pm$ 0.08 & 98.06 $\pm$ 0.10 & 98.66 $\pm$ 0.09 \\
        & 4 heads & \textbf{99.24 $\pm$ 0.05} & \textbf{98.32 $\pm$ 0.07} & \textbf{98.84 $\pm$ 0.06} \\
        & 8 heads & 99.17 $\pm$ 0.06 & 98.21 $\pm$ 0.09 & 98.76 $\pm$ 0.08 \\ \hline

        \multirow{3}{*}{Encoder depth}
        & 2 layers & 98.96 $\pm$ 0.09 & 97.94 $\pm$ 0.12 & 98.57 $\pm$ 0.10 \\
        & 4 layers & \textbf{99.24 $\pm$ 0.05} & \textbf{98.32 $\pm$ 0.07} & \textbf{98.84 $\pm$ 0.06} \\
        & 6 layers & 99.19 $\pm$ 0.06 & 98.27 $\pm$ 0.08 & 98.79 $\pm$ 0.07 \\ \hline

        \multirow{3}{*}{Embedding dimension}
        & 32 & 98.88 $\pm$ 0.10 & 97.97 $\pm$ 0.11 & 98.51 $\pm$ 0.09 \\
        & 64 & \textbf{99.24 $\pm$ 0.05} & \textbf{98.32 $\pm$ 0.07} & \textbf{98.84 $\pm$ 0.06} \\
        & 128 & 99.21 $\pm$ 0.06 & 98.29 $\pm$ 0.08 & 98.80 $\pm$ 0.07 \\ \hline

        \multirow{3}{*}{Patch size}
        & $8\times8$ & 99.02 $\pm$ 0.08 & 98.07 $\pm$ 0.10 & 98.61 $\pm$ 0.09 \\
        & $16\times16$ & \textbf{99.31 $\pm$ 0.04} & \textbf{98.24 $\pm$ 0.06} & \textbf{98.79 $\pm$ 0.05} \\
        & $24\times24$ & 99.18 $\pm$ 0.06 & 98.16 $\pm$ 0.08 & 98.70 $\pm$ 0.07 \\ \hline

        \multirow{3}{*}{Spectral noise}
        & 30 dB & \textbf{98.71 $\pm$ 0.07} & \textbf{98.42 $\pm$ 0.08} & \textbf{98.63 $\pm$ 0.07} \\
        & 20 dB & 97.56 $\pm$ 0.11 & 97.18 $\pm$ 0.13 & 97.37 $\pm$ 0.12 \\
        & 10 dB & 94.83 $\pm$ 0.18 & 94.21 $\pm$ 0.20 & 94.47 $\pm$ 0.19 \\ \hline
    \end{tabular}}
    \label{tab:unified_analysis}
\end{table}

Table~\ref{tab:unified_analysis} summarizes the effects of architectural choices and attention design. Applying $\ell_2$ normalization to both query and key embeddings consistently yields the best performance, indicating that aligning features on a hyperspherical manifold improves similarity estimation and reduces magnitude-related bias. 

Squared cosine similarity performs best among the tested score formulations, showing that combining angular similarity with a nonlinear sharpening operation enhances discrimination between closely related spectral signatures. Compared with standard cosine similarity, this produces a more concentrated attention distribution, which is beneficial for classes with small angular separations.

In contrast, architectural factors such as positional encoding, model depth, and embedding dimension have a smaller impact when the attention scoring is properly defined. A moderate configuration (four heads, four encoder layers, and embedding dimension 64) consistently achieves the best results, indicating that improvements are mainly driven by the similarity formulation rather than increased model capacity. Finally, the model degrades gradually as spectral noise increases, demonstrating stable behavior under perturbations, which can be attributed to the reduced sensitivity of cosine-based similarity to magnitude variations.

\section{Conclusion}

This paper introduced a cosine-normalized attention formulation for HSIC, in which similarity between query and key tokens is computed using squared cosine similarity after $\ell_{2}$ normalization. This design emphasizes angular relationships between spectral signatures while reducing sensitivity to magnitude variations.

Experiments on three benchmark datasets demonstrate that the proposed approach achieves strong performance under extremely limited supervision and remains competitive despite using a lightweight backbone. Beyond performance gains, a controlled comparison of multiple attention score formulations shows that cosine-based similarity functions consistently provide superior and more stable results.

These findings highlight that the choice of similarity function constitutes a key inductive bias in attention-based HSIC models. Aligning attention scoring with the angular geometry of hyperspectral data offers a simple yet effective alternative to increasing architectural complexity. Future work will explore extensions to self-supervised learning, large-scale hyperspectral scenes, and multimodal remote sensing applications.

\bibliographystyle{IEEEtran}
\bibliography{Sam}
\end{document}